\RequirePackage{iftex}
\ifPDFTeX \pdfoutput=1 \fi 
\documentclass[preprint,12pt]{elsarticle}

\usepackage[T1]{fontenc}
\usepackage[utf8]{inputenc}
\usepackage{lmodern}
\usepackage{textcomp}

\usepackage{amsmath,amssymb}

\usepackage{newunicodechar}
\newunicodechar{β}{\ensuremath{\beta}}
\newunicodechar{Δ}{\ensuremath{\Delta}}
\newunicodechar{μ}{\ensuremath{\mu}}
\newunicodechar{τ}{\ensuremath{\tau}}
\newunicodechar{ϵ}{\ensuremath{\epsilon}}
\newunicodechar{∝}{\ensuremath{\propto}}
\newunicodechar{∣}{\ensuremath{\mid}}
\newunicodechar{×}{\ensuremath{\times}}
\newunicodechar{−}{\ensuremath{-}}
\newunicodechar{≥}{\ensuremath{\geq}}
\newunicodechar{≤}{\ensuremath{\leq}}
\newunicodechar{↔}{\ensuremath{\leftrightarrow}}

\newunicodechar{š}{\v{s}}
\newunicodechar{č}{\v{c}}
\newunicodechar{ž}{\v{z}}
\newunicodechar{ś}{\'{s}}
\newunicodechar{ć}{\'{c}}
\newunicodechar{ż}{\.{z}}
\newunicodechar{ł}{\l}
\newunicodechar{Ł}{\L}
\newunicodechar{ę}{\k{e}}

\usepackage{longtable,booktabs,array}
\usepackage{multirow}
\usepackage{calc}
\usepackage{etoolbox}
\makeatletter
\patchcmd\longtable{\par}{\if@noskipsec\mbox{}\fi\par}{}{}
\makeatother

\usepackage{graphicx}
\graphicspath{{figures/}}
\usepackage{enumitem}

\usepackage{hyperref}
\IfFileExists{xurl.sty}{\usepackage{xurl}}{}
\hypersetup{colorlinks=true,allcolors=[rgb]{0.0,0.0,0.55},breaklinks=true}

\journal{Computer Speech \& Language}

\begin{document}

\begin{frontmatter}

\title{The Dynamics of Human and AI-Generated Language --- How Semantics
Fluctuates across Different Timescales\texorpdfstring{\tnoteref{lic}}{}}
\tnotetext[lic]{\copyright\,2026. This manuscript version is made available under the
CC\,BY-NC-ND~4.0 license (\url{https://creativecommons.org/licenses/by-nc-nd/4.0/}).
Formally published article: \url{https://doi.org/10.1016/j.csl.2026.102013}.}

\author{Han-Jen Chang}
\author{Yasir Çatal}
\author{Angelika Wolman}
\author{Agustín Ibáñez}
\author{David Smith}
\author{I-Wen Su}
\author{Kai-Yuan Cheng}
\author{Georg Northoff}

\date{April 30, 2026}

\begin{abstract}
Spoken language, whether produced by humans or large language models (LLM), unfolds over time with varying semantic content. However, we still lack simple, interpretable time-series features that capture how generic versus specific content is distributed over time, and that can be used to compare human and AI-generated speech. We introduce a semantic-timescale analysis pipeline that turns word-level transcripts with timestamps into semantic time-series. For each spoken narrative, we compute (i) semantic specificity using WordNet-based word depth and (ii) contextual similarity using SBERT embeddings and quantify their temporal dependence using autocorrelation-window measures (ACW-0 and related metrics). We then compare original speech to multiple shuffled controls that selectively disrupt lexical identity, temporal order, and word duration. Across human-read autobiographical narratives, TTS readings, and LLM-generated texts rendered with TTS, we find that segments with longer ACW-0 in the semantic time-series tend to contain more generic vocabulary, whereas segments with shorter ACW-0 are enriched in more specific words. These associations are strongly attenuated or abolished when word order and timing are randomized, indicating that ACW-based measures capture non-trivial temporal organization of semantic content beyond static lexical distributions. Our results suggest that ACW-based semantic timescales are a useful family of features for analyzing and comparing the temporal structure of human and AI-generated speech.
\end{abstract}


\end{frontmatter}

\section*{Introduction}

Timescales are a central concept across dynamic systems, from enzyme kinetics and ecosystems to neural activity and language (Çatal et al., 2024; Henzler-Wildman et al., 2007; Regev et al., 2024; Scheffer \& Carpenter, 2003). In each case, timescales index how long meaningful events last or at what rate information is integrated. In neuroscience, a hierarchy of intrinsic timescales supports the processing of sensory information over different temporal windows: shorter timescales in early sensory areas track rapidly changing inputs, whereas longer timescales in higher-order areas integrate over sentences and paragraphs (Hasson et al., 2008; Honey et al., 2012; Lerner et al., 2011; Wolff et al., 2022). Language is one of the domains where this logic applies most directly: information unfolds over time, and the temporal scale at which it is integrated --- from phonemes and words to sentences, discourses, and extended narratives --- has been a central organizing concept across psycholinguistics, speech science, and neuroscience (Levelt, 1989; Hasson et al., 2015).

These traditions have approached the relationship between information and time from different angles, each illuminating a different temporal scale. From an information-theoretic perspective, the entropy rate constancy principle and the uniform information density (UID) hypothesis have characterized how information is distributed across linguistic units (Genzel and Charniak, 2002; Jaeger and Levy, 2007). More recent work has shown that information content fluctuates in structured, predictable ways, forming "information contours" that are systematically related to discourse position, hierarchical structure, and grounding context (Giulianelli and Fernández, 2021; Giulianelli et al., 2021; Tsipidi et al., 2024; Gay et al., 2026), exhibiting spectral and periodic organization (Yang et al., 2023; Xu et al., 2024; Tsipidi et al., 2025; Ou et al., 2025), and converging over the course of dialogue (Xu and Reitter, 2018). In parallel, research has established that information and timing are coupled at the realization level: more predictable words are produced with shorter durations and reduced acoustic detail, linking surprisal to fine-grained prosodic and articulatory structure (Aylett, 1999; Aylett and Turk, 2004, 2006; Bell et al., 2003).

Across these traditions, however, time enters the analysis in only two forms. In the first, time is the ordinal position of discrete units (tokens, sentences, paragraphs) along a text sequence, with no physical duration attached. In the second, time appears as the duration of an individual speech unit, treated as a per-unit scalar dependent variable to be explained by its predictability --- yielding one data point per word or syllable. What has received comparatively little attention is a third treatment, in which time is neither a position index nor a per-unit scalar, but the continuous axis along which a semantic signal unfolds --- where the object of analysis is not the duration of any individual word, but the temporal pattern that emerges when semantic values are laid out in wall-clock time across multi-second windows, characterized directly with time-series measures such as autocorrelation and spectral decomposition.

This motivates a direct question: when semantic content itself is treated as a signal aligned to actual spoken time, does it exhibit this kind of non-random temporal organization? And does this organization systematically track properties of the content, such as how generic or specific it is? The present work addresses these questions using timescale tools drawn from the study of dynamical systems (e.g., autocorrelation-window measures), complementing rather than replicating the surprisal-based analyses.

To address these gaps, we rearrange and extend existing methods into a unified framework for semantic timescale analysis of speech. First, we convert continuous speech---traditionally analyzed through acoustic features---into high-resolution semantic time series by combining WordNet-based word depth as a proxy for lexical specificity (Miller, 1995; Princeton University, 2010) with SBERT embeddings to capture contextual similarity (Reimers \& Gurevych, 2019). Second, we characterize the temporal structure of these signals by bringing together complementary tools from two traditions: the autocorrelation window' s first zero-crossing (ACW-0), a time-domain measure of sustained temporal dependency developed in the study of intrinsic neural timescales (Honey et al., 2012; Golesorkhi et al., 2021; Wolff et al., 2022), and frequency-domain descriptors --- the power-law exponent (PLE) and mean frequency (MF) --- analogous to the spectral tools that have recently been used to characterize information contours in text (Yang et al., 2023; Tsipidi et al., 2025; Ou et al., 2025). Together, these yield a joint time--frequency description of how specificity and contextual meaning fluctuate over real spoken time. Third, we construct a battery of shuffled control conditions that selectively disrupt lexical identity, temporal order, and word--duration pairings, allowing us to test whether the observed timescales reflect trivial properties of speech timing or non-random temporal organization of semantic content.

We investigate the generic--specific axis because it offers a simple, theory-neutral operationalization with clear implications for language processing. Behavioral and neuroimaging works suggest that specificity engages partially distinct processing routes, with more specific concepts often associated with higher processing demands and different activation patterns than more generic concepts (Paivio, 1991; Crossley et al., 2009, 2012; Binder et al., 2016; Bolognesi et al., 2020; Bi, 2021; Vignali, 2023; Jamali et al., 2024). Although concreteness and specificity are not identical, they tend to be moderately correlated (Crossley et al., 2009, 2012), making the generic--specific gradient a useful low-dimensional axis to probe semantic organization. Building on these observations, we hypothesize that the temporal organization of speech will differ as a function of its semantic content: segments enriched in generic, low-depth words will exhibit longer ACW-0, consistent with broader, more sustained semantic spans, whereas segments enriched in specific, high-depth words will exhibit shorter ACW-0, consistent with more locally concentrated semantic content. This prediction is in line with the broader principle that different kinds of content are integrated over different characteristic timescales, a pattern well documented in the neural processing of language (Lerner et al., 2011; Hasson et al., 2015).

Within this framework, we ask three related questions. First (Q1), do semantic signals extracted from speech---operationalized as WordNet depth and SBERT-based semantic similarity time series---exhibit non-random temporal structure when compared with shuffled or temporally randomized controls? Second (Q2), is there a systematic relationship between the length of semantic timescales (ACW-0) and lexical specificity, such that segments dominated by generic versus specific content occupy different temporal windows, and do these relationships survive when we selectively disrupt word identity, order, or timing? Third (Q3), are these patterns robust across human speech, human-derived text-to-speech (TTS), and LLM-derived TTS, or are they specific to one production modality? Framing the work in terms of these questions emphasizes that our goal is not to claim that timescales ``create'' meaning, but to provide operational, testable descriptors of how semantic information is temporally organized in speech.

In summary, we propose a general pipeline that turns continuous speech into semantic time series and characterizes their temporal structure using tools that have so far been applied mainly to neural data (e.g., ACW-0). By combining this pipeline with carefully designed shuffling controls, dynamic alignment procedures (e.g., dynamic time warping; Müller, 2007), and joint time--frequency characterization (ACW-0, PLE, MF), and by applying it to human narratives, conventional TTS, and LLM-derived TTS, we test whether semantic timescales are (i) present and non-random in spoken language, (ii) systematically related to generic versus specific lexical content, and (iii) applicable to both human and AI-generated speech. However, we treat the LLM-derived condition as an exploratory extension rather than as a clean test of human-versus-AI language representations, because the resulting timing is jointly shaped by generated content and TTS synthesis. In addition, recent work has highlighted limitations in current LLMs' ability to comprehensively capture meaning (Dentella et al., 2024). We therefore interpret these findings cautiously, as supporting the comparative usefulness of timescale-based semantic features across different speech-generation conditions rather than as evidence for equivalence between human and AI language processing. This opens a path toward integrating timescale-based semantic features into computational models of speech processing and into emerging clinical applications of large language models (e.g., coherence and thought-disorder assessment; Holmlund et al., 2023; Palominos et al., 2024).

\begin{figure}[htbp]
\centering
\includegraphics[width=\linewidth]{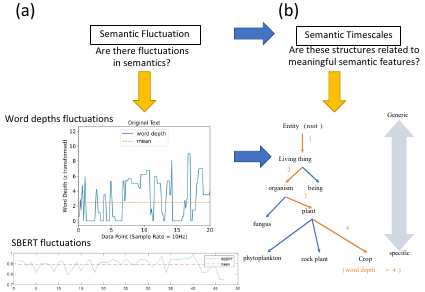}
\caption{Analytical pipeline overview. (a)~We investigate whether there are fluctuations and dynamics in semantics using word depths and SBERT similarity. (b)~Next, we ask whether specific semantic contents (word depths) are associated with longer or shorter timescales in these fluctuations.}
\label{fig:pipeline}
\end{figure}

\section{Methods}

\subsection{Subjects and Data Acquisition}

Three datasets were explored in this study: Human-Derived Human-Read(H-H), Human-Derived Text-to-Speech model-Read (H-TTS), and LLM-derived Text-to-Speech model-read (LLM-TTS). For the first dataset, twenty-seven healthy subjects (10 females; mean age = 30.3 years) were recruited from the local community in Ottawa, Canada. The experimental protocol was approved by the ethics commission (REB \#2016004) of the University of Ottawa Institute of Mental Health Research.

The recordings of spoken language were obtained from the study by Smith et al. (2022), where participants were requested to record unstructured, eight-minute autobiographical narratives (\emph{M} = 1103 words, \emph{SD} = 221 words). In our analyses, we further excluded three subjects due to low accuracy of text transcription following preprocessing. An additional subject was excluded due to an unsupported language in WordNet, leading to a resized sample size of 17 subjects.

We replicated all the analyses with two other datasets. The H-TTS dataset consists of audios generated by Text-to-Speech (TTS) model developed by OpenAI (OpenAI, 2024), which used the original texts transcribed from the self-introductions of the 17 H-H subjects. The LLM-TTS dataset comprises twenty audios converted by TTS based on GPT-4 generated texts (\emph{M} = 600 words, \emph{SD} = 67 words). The prompt used was ``you are a human participant of a research project. Please assign to yourself a name. And you are now asked to give an oral self-introduction with around 5000 words''. (See Supplementary Methods 1, 2 for details of TTS and speech-to-text transcriptions)

\subsection{Sampling of WordNet depths and SBERT similarity}

We derive two simple, one-number-per-moment signals from continuous speech.\\
(i) WordNet-depth (specificity) signal. Each spoken word is linked to its WordNet synsets (senses) (WordNet version:WordNet 3.0 via NLTK package 3.9.1, 2024). In the WordNet hierarchy, a synset's depth is the length of the shortest path from the root (``unique beginner'') to that synset (Fig. 2c). Because many words are polysemous, we define the word's depth as the mean synset depth (lower depth = more generic; higher depth = more specific). We emphasize that this measure is a coarse lexical-taxonomic proxy for specificity rather than a full context-sensitive representation of word meaning. WordNet organizes words into synsets linked by semantic relations, but averaging across all synsets does not perform token-level word-sense disambiguation (WSD) and may therefore blur contextually distinct meanings. Our aim is therefore not to model the exact intended sense of each word, but to test whether this coarse lexical-specificity signal, once aligned to real spoken time, exhibits non-random temporal organization across discourse. To assess whether sense-averaging substantially affected the signal, we additionally performed a WSD-based sensitivity check using the Lesk algorithm. The averaged word depths displayed high correlation (\(r = 0.8379,p < 0.001\)) with the WSD-based word depths (Supplementary Result 9). (Fig. 2a).

To assess whether WordNet depth reflects meaningful lexical-semantic structure in our dataset, we correlated it with independent lexical norms on the human speech vocabulary, including specificity ratings (Muraki and Pexman, 2026) and concreteness ratings (Brysbaert et al., 2014). WordNet depth was positively correlated with both concreteness (\emph{r} = 0.460, \emph{p} \ensuremath{<} 0.001) and specificity (\emph{r} = 0.143, \emph{p} \ensuremath{<} 0.001). Consistent with prior work distinguishing concreteness from categorical specificity and relating WordNet-like hierarchies to abstraction/specificity (e.g., Bolognesi et al., 2020; Puccetti et al., 2025; Ravelli et al., 2025), we retain WordNet depth here as a coarse, taxonomy-based proxy of specificity.

(ii) SBERT-based similarity signal. We segment the transcript into overlapping 100-s windows with a 10-s step (Fig. 2a--b). Each window is embedded with Sentence-BERT (SBERT), and we compute the cosine similarity between adjacent windows to obtain a bounded [−1,1] time series. Higher values mean the current snippet is semantically closer to the just-previous snippet; lower values mean content has shifted. This yields a running measure of semantic continuity through time.

\subsubsection*{Sampling and preprocessing}

For the WordNet signal, we sample at 10 Hz: every 0.1 s we assign the depth of the word(s) active at that instant; if multiple words fall within a frame, we average their depths. We exclude stop words (little lexical specificity in WordNet) and omit silent pauses rather than assigning semantic values to non-speech. The remaining depth values are concatenated into the specificity series. For the SBERT signal, adjacent-window cosine similarities produce the coarser-grained continuity series aligned to the 10-s step. Each segment was transformed into vectors with SBERT (SBERT: Sentence-BERT for Sentence Embeddings. (2022). GitHub. Retrieved July 15, 2022, from \url{https://huggingface.co/sentence-transformers/all-MiniLM-L6-v2}, model: sentence-transformers/all-MiniLM-L6-v2). Cosine similarity was then measured between vectors of every two adjacent timeframes.

\begin{figure}[htbp]
\centering
\includegraphics[width=\linewidth]{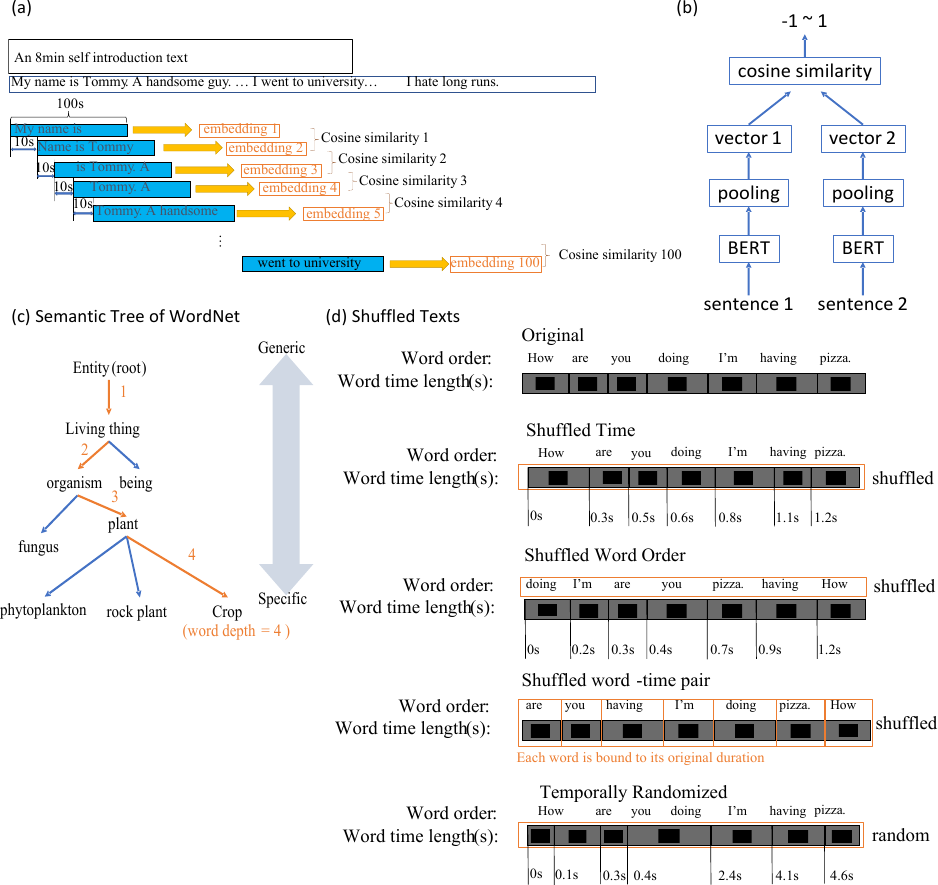}
\caption{(a) Text is cut into multiple overlapping 100s timeframes with 10s interval to examine the fluctuation of SBERT. (b) SBERT's procedure of calculating sentence similarity (Reimers and Gurevych, 2019, p.3) (c) Semantic tree. (Pawar et al., 2018, p4) An example of how word depth is calculated. In the case, the word depth of ``crop'' is measured by its closest distance to its root ``entity'', which is 4. (d) Four different groups are tested to illustrate the dynamics of word depth and SBERT. The same data of one subject is used to generate the shuffled time and shuffled word order texts. In the shuffled time texts, the time length of each word is calculated by subtracting the time onset of the current word from the onset of the next word. Then the time length of each word is shuffled and reassigned to the words randomly. In the shuffled word order texts, only the word order is shuffled while the onset of each word block remains the same. The shuffled word-time pair consists of shuffled segments where the word and its duration are tied together. The texts of temporally randomized group compose of texts with original word order but duration of words is randomly assigned ranging from 0.1s to 2.0s.}
\label{fig:methods}
\end{figure}

\subsection{Shuffled texts}

Original text was used to generate different shuffled texts (as shown in Figure 2d).

\textbf{Shuffled Time (ST):} In this condition, the original sequence of words was preserved, but the real-world durations associated with each word were randomly permuted among them. This control tests whether the observed timescales are merely a product of the distribution of word durations, irrespective of which word is spoken for how long.

\textbf{Shuffled Word Order (SWO):} This condition preserved the exact temporal scaffolding of the original speech---maintaining each word' s original onset time and duration---but randomly permuted the sequence of words. By isolating the influence of linguistic content from its timing, this control is critical for testing whether the specific sequence of semantic concepts contributes to the temporal structure.

\textbf{Shuffled Word-Time Pair (SWTP):} Here, each word and its corresponding duration were kept together as a pair, but the sequence of these word-duration pairs was randomly shuffled. This control tests whether the observed timescales are an emergent property of the larger narrative sequence, rather than just the statistical distribution of word-duration pairings.

\textbf{Temporally Randomized (TR):} This control maintained the original word order but replaced the natural, human-produced word durations with durations randomly assigned from a uniform distribution ranging from 0.1s to 2.0s. This allows us to test whether the specific, nuanced timing of human speech is critical for generating the observed timescales, as opposed to any arbitrary timing.

Together, these four controls allow us to disentangle trivial temporal properties from genuinely structured semantic dynamics. Shuffled durations (ST) test whether effects can be explained by the marginal distribution of word lengths and speaking rate; shuffled orders (SWO) isolate the impact of lexical sequencing given intact timing; shuffled word--time pairs (SWTP) randomize larger semantic-prosodic chunks; and temporally randomized speech (TR) removes realistic timing altogether. Any effect that survives in the original texts but disappears or reverses under these manipulations cannot be attributed to simple differences in average word duration or segment length.

\subsection{Comparing fluctuations of Original and Shuffled texts.}

To determine whether the time-domain and frequency-domain of the fluctuations change after shuffling time and word order, we compared the fluctuations of original texts with the shuffled texts using dynamic time warping (DTW), and the differences in autocorrelation window-0 (ACW-0), mean frequency of PSD and PLE. (See Methods 1.5., 1.6., 1.7. for more details of each measurement) The word depth and SBERT were acquired as in Method 1.2.

To assess whether shuffling alters text structure, we employed a pseudo-original surrogate Monte Carlo test (Theiler \& Prichard, 1996), a form of surrogate data testing closely related to permutation testing. For each subject, the effect size was defined as

\[\Delta_{\mathrm{on}}\, = \, m_{\text{orig}}\, - \,\frac{1}{K}\sum_{i = 1}^{K}m_{\mathrm{shuff}}\,\]

where \(m_{\mathrm{shuff}}\) denotes the metric (ACW-0, PSD, PLE). The null hypothesis stated that the original and shuffled versions are exchangeable, i.e., \(\Delta_{\mathrm{on}}\) is indistinguishable from the variability among shuffled texts. In each Monte Carlo replicate (B = 10,000), one version (original or shuffled) was randomly designated as the ``pseudo-original'' among the 200 shuffled versions and its difference from the mean of the remaining versions (\(\Delta^{*}\)) was recorded to generate a null distribution. The observed \(\Delta_{\mathrm{on}}\) was then compared to this distribution of \(\Delta^{*}\) to obtain per-subject p-values.

Due to heavier computation requirements for DTW, a permutation test with 100 rounds of shuffled texts were adopted to test for effects of shuffling on DTW. The Sakoe-Chiba band was set to 5\%. From the resulting \((K{+}1)\times(K{+}1)\) distances (with K=100), we computed, for each sequence i, the mean DTW distance to all other sequences, yielding test statistics \(\{M_0, M_1, \ldots, M_K\}\), where \(M_0\) corresponds to the original. Under the null hypothesis that any sequence could be the ``original,'' we treat all K+1 cases as exchangeable. The p-value of permutation test is calculated by \(p = \frac{1 + \#\{\, i:\left| M_{i} - \mu \right| \geq \left| M_{0} - \mu \right|\,\}}{(K + 1) + 1}\) where μ is the mean of \(\{M_i\}\) and the ``+1'' correction ensures non-zero p-values in accordance with standard empirical permutation testing practice.

To combine p-values across subjects while accounting for possible covariances among subjects introduced by shared processing pipelines, we used the Empirical Brown's Method (EBM) (Brown, 1975; Poole et al., 2016), which provides more reliable Type I errors.

\subsection{Dynamic time warping (DTW)}

Dynamic time warping (DTW) (dtaidistance, 2023) outputs the distance to the optimal alignment between two signals. Generally, a higher distance in DTW indicates higher dissimilarity between two sequences. DTW was used to compare the fluctuations of Word Depths and SBERT similarity of each subject between its original and shuffled versions.

\subsection{Power spectrum}

The fluctuations of word depths and SBERT similarities of each subject were converted into power spectra using the Signal Processing Toolbox in MATLAB (2022). The mean frequency of power spectra was calculated, and the power spectra were also transformed into the log-log format to estimate the slope (PLE) in linear regression. Both the mean frequencies and slopes were compared between original text and shuffled texts

\subsection{Autocorrelation and Autocorrelation window}

Autocorrelation window -- 0 (ACW-0) were calculated for the WD and SBERT fluctuations of each subject using the Statsmodels library (Seabold, S., \& Perktold, J., 2010) in Python (3.8).

An autocorrelation window (ACW) is the lag where the correlation value of an autocorrelation function, which measures the repeating patterns in a signal, decreases to a certain value. The ACW-0 variant is the lag that the correlation first decreases to 0. Since the correlation values often do not show lags on exact zero, we determined ACW-0 as the lag where the ACF decreases to lower than zero or the lag before it, depending on whose absolute distance to zero is smaller (Golesorklhi et al., 2021).

Other timescale metrics can also characterize signal dynamics, including full width at half maximum (FWHM), which quantifies the width of the ACF above 0.5 (Honey et al, 2012), and the e-folding time (τ), which assumes exponential decay and estimates the lag at which the ACF falls to 1/e (Murray, 2014; \href{https://www.cell.com/neuron/fulltext/S0896-6273(15)00765-5?_returnURL=https\%3A\%2F\%2Flinkinghub.elsevier.com\%2Fretrieve\%2Fpii\%2FS0896627315007655\%3Fshowall\%3Dtrue}{Chaudhuri}, 2015). We adopt ACW-0 as our primary measure because it is model-agnostic, imposes minimal assumptions about ACF shape, and aligns with our exploratory goal of testing whether language exhibits seconds-scale temporal organization. Its definition is also operationally straightforward, which is the lag required for complete loss of periodicity and memory, providing a clear and interpretable descriptor of timescale in our setting. We treat ACW-0---the first zero-crossing of the autocorrelation---as an indicator of persistence timescales: how long the fluctuation pattern tends to remain statistically similar to its recent past before decorrelating (Golesorkhi et al., 2021). Interpreted this way, a longer ACW-0 for the WordNet-depth series means the pattern of specificity changes (generic↔specific ups-and-downs) persists longer before it loses resemblance to earlier behavior; a shorter ACW-0 means quicker alternation. For the SBERT series, a longer ACW-0 means the pattern of local similarity vs. dissimilarity across time shows stronger serial dependence (not necessarily high similarity at every step), whereas a shorter ACW-0 indicates faster semantic drift.

For completeness, we also report ACW-50/40/30/20/10 as secondary analyses (Fig. 4b), akin to prior uses of FWHM (ACW-50).

\subsection{Relationship between Temporal windows (ACW-0) and Semantics (Word Depth of WordNet)}

To measure if there are certain relationships between ACW-0 and word depths, we sliced the fluctuations into segments and calculate whether the ACW-0 correlates with the average word depths within these segments. Only the WordNet specificity fluctuations were used in this analysis due to better representation of semantic content.

First, each subject's texts were cut into overlapping segments of 1000 datapoints with a step size of 100 datapoints (10\%). Since the datapoints were sampled at 10Hz, 1000 datapoints equate to 100s timeframe with 10s steps. The WD of each timeframe is the average of the 1000 WD datapoints.

Then, the ACW-0s and average word depth of all segments were calculated. To normalize the ACW-0 values for each subject and capture potential non-linear relationships, the unique values of ACW-0 of each subject were categorized into 3 categories based on their position within the sorted distribution: category 1 for the lower one-third, category 2 for the middle one-third, category 3 for the higher one-third. If a subject had fewer than three unique values, the values were categorized into one or two categories according to the number of unique values. To ensure consistency between ACW-0 categories and word depths, we normalized ACW-0 categories for each subject and multiplied them by the standard deviation of the subject's word depths to match the variation range of the two variables. (categorization into 4 categories revealed similar results, see supplement results 1)

A Linear Mixed Effect model (LMM) was then employed to calculate the relationship between word depths and categorized ACW-0. The LMM for the original texts is expressed as:

\[ \textit{Word Depth}_{ij} = \beta_0 + \beta_1\,\textit{ACW0c}_{ij} + \left(1 \mid \textit{Subject}_i\right) + \epsilon_{ij} \]

and for shuffled texts:

\[ \textit{Word Depth}_{ij} = \beta_0 + \beta_1\,\textit{ACW0c}_{ij} + \left(1 \mid \textit{Subject}_i\right) + \left(1 \mid \textit{Subject}_i\!:\!\textit{Round}_j\right) + \epsilon_{ij} \]

This model examines how changes in ACW-0 categories (\(\textit{ACW0c}\)) influence word depth. Here, \(\beta_0\) is the intercept, and \(\beta_1\) quantifies the fixed effect. To further account for the random effects based on subjects and the 200 rounds of shuffling, we incorporated random intercepts for individual subjects \((1 \mid \textit{Subject}_i)\) and for rounds that might differ across subjects \((1 \mid \textit{Subject}_i\!:\!\textit{Round}_j)\). \(\epsilon_{ij}\) represents the model's residual error unexplained by the previous variables. We compared the results with those of the shuffled texts and temporally randomized texts to further probe if such correlation only occurs in original texts.

As an additional sensitivity analysis, we re-estimated the LMMs with mean Zipf frequency included as a covariate to test whether the ACW--word-depth relationship could be explained by lexical frequency. The formula used was:

\[ \textit{Word Depth}_{ij} = \beta_0 + \beta_1\,\textit{ACW0c}_{ij} + \beta_2\,\textit{Zipf}_{ij} + \left(1 \mid \textit{Subject}_i\right) + \left(1 \mid \textit{Subject}_i\!:\!\textit{Round}_j\right) + \epsilon_{ij} \]

Zipf frequency was calculated with English Zipf frequency from the wordfreq package (wordfreq 3.1.1, Spear, 2022)

\subsection{Statistical analysis}

All analyses were performed in Python (version 3.8) with standard scientific libraries including NumPy, pandas, and SciPy. We implemented a pseudo-original surrogate Monte Carlo test (Theiler \& Prichard, 1996) to assess per-subject effects (Δ = original − mean shuffled), simulating the null distribution across 10,000 iterations and applying the Phipson--Smyth +1 correction to avoid zero p-values (Phipson \& Smyth, 2010). To test the global null hypothesis that no effect exists across all subjects, we combined the per-subject p-values using EBM. This approach yields a single group-level p-value, thus obviating the need for a separate multiple comparison correction such as FDR for this primary analysis.

In regards of the LMM, the residuals were checked for normal distribution with Kolmogorov-Smirnov test (Kolmogorov, 1933; Smirnov, 1948) due to larger sample sizes of ACW-0. Although all original texts exhibit normality, the shuffled texts, possibly due to large sample sizes (Mean = 140871.25, SD = 96116.22), did not pass the Kolmogorov-Smirnov test. Nonetheless, the histograms and Q-Q plots of shuffled texts conform to normality. (see supplement results 4) Additionally, similar results from Spearman correlations (Spearman, 1904) further support the results of LMM. (see supplement results 4) The LMM analyses were carried out in RStudio (Version 2024.04.2) and language R (Version 4.2.2).

\section{Results}

\subsection{Dynamic fluctuations of word depths and SBERT in time and frequency domains}

To investigate whether there are dynamic fluctuations in word depths and SBERT similarities that are susceptible to shuffling of time and word order, we compared the original texts with shuffled texts by measuring their deviation with dynamic time warping (DTW), and their differences in autocorrelation window-0 (ACW-0), mean frequency of power spectrum density (PSD), and power law exponent (PLE) (see methods 1.6., 1.7., 1.8. for more details). Then, utilizing surrogate and permutation tests, we tested the null hypothesis: The difference between original and shuffled texts is statistically indistinguishable from the differences observed among shuffled texts themselves, and that the original text and shuffled versions are exchangeable. In practical terms, a significant result rejects interchangeability and indicates that a temporal structure present in the original is disrupted by shuffling. These analyses were conducted on all three datasets, including the Human-Derived Human-Read(H-H), Human-Derived Text-to-Speech model-Read (H-TTS), and LLM-derived Text-to-Speech model-read (LLM-TTS) data.

Significant changes in both word depths and SBERT similarity fluctuations were observed across all datasets. In the H-H dataset, the mean absolute distances (as measured by Dynamic Time Warping/DTW) of the word depth and SBERT fluctuations between original and shuffled texts are larger than zero and rejected interchangeability. Next, we compared original and shuffled data based on their ACW-0 values. These results indicate significant differences of ACW-0 in word depth as well as in SBERT similarity fluctuation among original and shuffled texts. (see Table 2). We further verified that these SBERT-based findings were robust to window parameterization: replicating the analysis on the H-H dataset with a 50-s window and 5-s step preserved all four effects (DTW, ACW-0, mean frequency, PLE; all p \ensuremath{<} 0.016; Supplementary Result 10).

{\centering\small\textbf{Table 1.} Dynamic Time Warping comparing the fluctuations of different groups.\par}\nobreak\smallskip
\begin{longtable}[]{@{}
  >{\centering\arraybackslash}p{(\linewidth - 8\tabcolsep) * \real{0.1246}}
  >{\centering\arraybackslash}p{(\linewidth - 8\tabcolsep) * \real{0.1279}}
  >{\centering\arraybackslash}p{(\linewidth - 8\tabcolsep) * \real{0.1503}}
  >{\centering\arraybackslash}p{(\linewidth - 8\tabcolsep) * \real{0.2216}}
  >{\centering\arraybackslash}p{(\linewidth - 8\tabcolsep) * \real{0.2308}}@{}}
\toprule\noalign{}
\begin{minipage}[b]{\linewidth}\centering
Dataset
\end{minipage} & \begin{minipage}[b]{\linewidth}\centering
Score type
\end{minipage} & \begin{minipage}[b]{\linewidth}\centering
Group type
\end{minipage} & \begin{minipage}[b]{\linewidth}\centering
\(\overline{\mathbf{|Distance(dtw)|}}\)
\end{minipage} & \begin{minipage}[b]{\linewidth}\centering
P
\end{minipage} \\
\midrule\noalign{}
\endhead
\bottomrule\noalign{}
\endlastfoot
\multirow{8}{=}{\centering\arraybackslash H-H} & \multirow{4}{=}{\centering\arraybackslash WD} & Orig. vs. ST & 108.207 (40.678) & \ensuremath{<}0.001 \\
& & Orig. vs. SWO & 121.439 (35.155) & \ensuremath{<}0.001 \\
& & Orig. vs. SWTP & 79.267 (4.676) & 0.001 \\
& & Orig. vs. TR & 367.729 (100.025) & \ensuremath{<}0.001 \\
& \multirow{4}{=}{\centering\arraybackslash SBERT} & Orig. vs. ST & 0.395 (0.202) & \ensuremath{<}0.001 \\
& & Orig. vs. SWO & 0.323 (0.060) & \ensuremath{<}0.001 \\
& & Orig. vs. SWTP & 0.434 (0.171) & \ensuremath{<}0.001 \\
& & Orig. vs. TR & 4.375 (2.519) & \ensuremath{<}0.001 \\
\multirow{8}{=}{\centering\arraybackslash H-TTS} & \multirow{4}{=}{\centering\arraybackslash WD} & Orig. vs. ST & 49.147 (15.179) & \ensuremath{<}0.001 \\
& & Orig. vs. SWO & 76.669 (8.598) & \ensuremath{<}0.001 \\
& & Orig. vs. SWTP & 71.382 (6.896) & \ensuremath{<}0.001 \\
& & Orig. vs. TR & 372.680 (93.758) & \ensuremath{<}0.001 \\
& \multirow{4}{=}{\centering\arraybackslash SBERT} & Orig. vs. ST & 0.347 (0.155) & \ensuremath{<}0.001 \\
& & Orig. vs. SWO & 0.317 (0.039) & \ensuremath{<}0.001 \\
& & Orig. vs. SWTP & 0.406 (0.124) & \ensuremath{<}0.001 \\
& & Orig. vs. TR & 3.978 (2.355) & \ensuremath{<}0.001 \\
\multirow{8}{=}{\centering\arraybackslash LLM-TTS} & \multirow{4}{=}{\centering\arraybackslash WD} & Orig. vs. ST & 35.778 (5.809) & \ensuremath{<}0.001 \\
& & Orig. vs. SWO & 61.223 (4.149) & \ensuremath{<}0.001 \\
& & Orig. vs. SWTP & 59.481 (4.063) & \ensuremath{<}0.001 \\
& & Orig. vs. TR & 257.652 (51.630) & \ensuremath{<}0.001 \\
& \multirow{4}{=}{\centering\arraybackslash SBERT} & Orig. vs. ST & 0.675 (0.325) & \ensuremath{<}0.001 \\
& & Orig. vs. SWO & 0.297 (0.055) & \ensuremath{<}0.001 \\
& & Orig. vs. SWTP & 0.628 (0.302) & \ensuremath{<}0.001 \\
& & Orig. vs. TR & 2.108 (1.118) & \ensuremath{<}0.001 \\
\end{longtable}

{\footnotesize\itshape Abbreviations: \emph{P}, combined p-value with Empirical Brown's test (EBM); \emph{SD}, standard deviation; \emph{Orig.}, original; \emph{ST}, shuffled time; \emph{SWO}, shuffled word order; \emph{SWTP}, shuffled word-time pair; \emph{TR}, temporally randomized.\par}

Together, these results show significant differences of original and shuffled data in both their time course (DTW) and time windows (ACW-0). Similarly, we compared the frequency domain by calculating the mean frequency of PSD and PLE, of which the differences all rejected interchangeability between original and shuffled texts. (Table 3 and Figure 3) This result suggests the presence of a certain temporal structure in the semantics of spoken language.

{\centering\small\textbf{Table 2.} Comparison of ACW-0 between original and shuffled texts.\par}\nobreak\smallskip
\begin{longtable}[]{@{}
  >{\centering\arraybackslash}p{(\linewidth - 10\tabcolsep) * \real{0.1755}}
  >{\centering\arraybackslash}p{(\linewidth - 10\tabcolsep) * \real{0.1369}}
  >{\centering\arraybackslash}p{(\linewidth - 10\tabcolsep) * \real{0.1420}}
  >{\centering\arraybackslash}p{(\linewidth - 10\tabcolsep) * \real{0.1889}}
  >{\centering\arraybackslash}p{(\linewidth - 10\tabcolsep) * \real{0.1177}}
  >{\centering\arraybackslash}p{(\linewidth - 10\tabcolsep) * \real{0.1060}}@{}}
\toprule\noalign{}
\begin{minipage}[b]{\linewidth}\centering
Dataset
\end{minipage} & \begin{minipage}[b]{\linewidth}\centering
Score type
\end{minipage} & \begin{minipage}[b]{\linewidth}\centering
Group type
\end{minipage} & \begin{minipage}[b]{\linewidth}\centering
$\overline{\mathbf{|\Delta ACW}\mathbf{0|}}$

(SD)
\end{minipage} & \begin{minipage}[b]{\linewidth}\centering
P
\end{minipage} & \begin{minipage}[b]{\linewidth}\centering
median \(\mathbf{\Delta}\)
\end{minipage} \\
\midrule\noalign{}
\endfirsthead
\toprule\noalign{}
\begin{minipage}[b]{\linewidth}\centering
Dataset
\end{minipage} & \begin{minipage}[b]{\linewidth}\centering
Score type
\end{minipage} & \begin{minipage}[b]{\linewidth}\centering
Group type
\end{minipage} & \begin{minipage}[b]{\linewidth}\centering
$\overline{\mathbf{|\Delta ACW}\mathbf{0|}}$

(SD)
\end{minipage} & \begin{minipage}[b]{\linewidth}\centering
P
\end{minipage} & \begin{minipage}[b]{\linewidth}\centering
median \(\mathbf{\Delta}\)
\end{minipage} \\
\midrule\noalign{}
\endhead
\bottomrule\noalign{}
\endlastfoot
\multirow{8}{=}{\centering\arraybackslash H-H} & \multirow{4}{=}{\centering\arraybackslash WD} & Orig. vs. ST & 11.88(12.85) & 0.004 & -8.48 \\
& & Orig. vs. SWO & 12.4(12.63) & 0.004 & -9.95 \\
& & Orig. vs. SWTP & 5.59(6.91) & 0.013 & -4.07 \\
& & Orig. vs. TR & 15.46(20.00) & 0.002 & -10.87 \\
& \multirow{4}{=}{\centering\arraybackslash SBERT} & Orig. vs. ST & 1.02(0.84) & 0.012 & -0.24 \\
& & Orig. vs. SWO & 1.24(0.92) & 0.005 & 1.015 \\
& & Orig. vs. SWTP & 1.33(0.93) & 0.004 & 0.985 \\
& & Orig. vs. TR & 2.03(1.12) & 0.001 & 1.995 \\
\multirow{8}{=}{\centering\arraybackslash H-TTS} & \multirow{4}{=}{\centering\arraybackslash WD} & Orig. vs. ST & 5.97(7.25) & 0.008 & -1.145 \\
& & Orig. vs. SWO & 7.58(7.43) & 0.006 & -4.745 \\
& & Orig. vs. SWTP & 6.18(6.16) & 0.007 & -3.58 \\
& & Orig. vs. TR & 12.67(13.49) & 0.002 & -11.525 \\
& \multirow{4}{=}{\centering\arraybackslash SBERT} & Orig. vs. ST & 1.16(1.36) & 0.013 & 0.12 \\
& & Orig. vs. SWO & 1.89(1.72) & 0.009 & -1.235 \\
& & Orig. vs. SWTP & 1.91(1.74) & 0.01 & -0.915 \\
& & Orig. vs. TR & 1.84(1.72) & 0.007 & -0.09 \\
\multirow{8}{=}{\centering\arraybackslash LLM-TTS} & \multirow{4}{=}{\centering\arraybackslash WD} & Orig. vs. ST & 6.08(9.32) & 0.012 & -1.532 \\
& & Orig. vs. SWO & 8.49(9.77) & 0.004 & -4.795 \\
& & Orig. vs. SWTP & 8.63(9.96) & 0.003 & -4.978 \\
& & Orig. vs. TR & 15.19(15.44) & 0.001 & -11.76 \\
& \multirow{4}{=}{\centering\arraybackslash SBERT} & Orig. vs. ST & 1.20(1.35) & 0.008 & 0.19 \\
& & Orig. vs. SWO & 1.80(1.27) & 0.005 & -1.53 \\
& & Orig. vs. SWTP & 1.64(1.15) & 0.004 & -0.885 \\
& & Orig. vs. TR & 1.99(1.81) & 0.006 & -0.522 \\
\end{longtable}

{\footnotesize\itshape Abbreviations: \emph{P}, combined p-value from EBM; \emph{SD}, standard deviation; \emph{Orig.}, original; \emph{ST}, shuffled time; \emph{SWO}, shuffled word order; \emph{SWTP}, shuffled word-time pair; \emph{TR}, temporally randomized.\par}

Beside the results of H-H dataset, the H-TTS and LLM-TTS datasets also presented with similar results. In these two datasets, the differences between all shuffled texts and original texts were significant as demonstrated by ACW-0, DTW, and PLE of both the fluctuations of WD and SBERT.

Together, complementing the results in the time domain, we show significant differences of original and shuffled data in the frequency domain that could not be produced by pure shuffled texts. This suggests the presence of a certain temporal structure with potentially semantically meaningful fluctuations in the spoken language.

\begin{figure}[htbp]
\centering
\includegraphics[width=\linewidth]{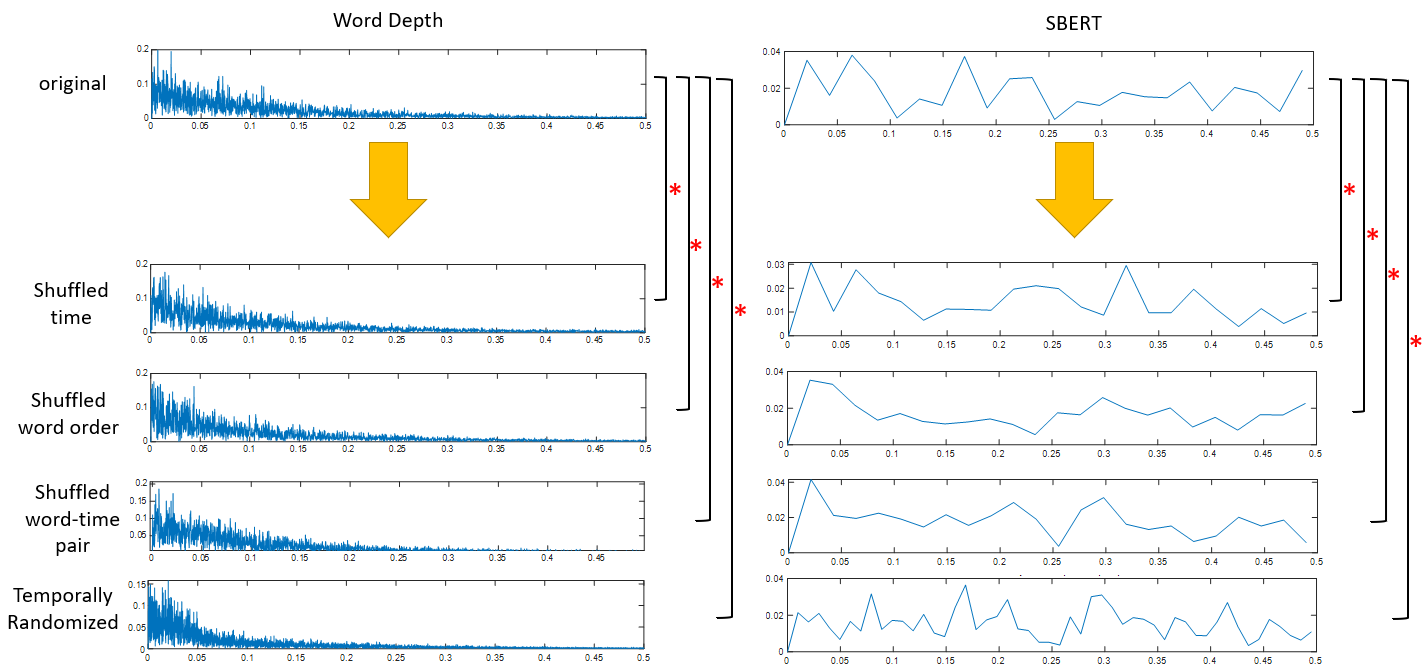}
\caption{Demonstration of one subject's PSD of Word Depths and SBERT similarities compared to the corresponding shuffled texts.}
\label{fig:psd}
\end{figure}

{\centering\small\textbf{Table 3.} Comparison of PLE and mean frequencies between original and shuffled texts.\par}\nobreak\smallskip
\begin{longtable}[]{@{}
  >{\centering\arraybackslash}p{(\linewidth - 16\tabcolsep) * \real{0.1128}}
  >{\centering\arraybackslash}p{(\linewidth - 16\tabcolsep) * \real{0.0986}}
  >{\centering\arraybackslash}p{(\linewidth - 16\tabcolsep) * \real{0.1018}}
  >{\centering\arraybackslash}p{(\linewidth - 16\tabcolsep) * \real{0.2080}}
  >{\centering\arraybackslash}p{(\linewidth - 16\tabcolsep) * \real{0.0845}}
  >{\centering\arraybackslash}p{(\linewidth - 16\tabcolsep) * \real{0.1127}}
  >{\centering\arraybackslash}p{(\linewidth - 16\tabcolsep) * \real{0.0836}}
  >{\centering\arraybackslash}p{(\linewidth - 16\tabcolsep) * \real{0.0854}}
  >{\centering\arraybackslash}p{(\linewidth - 16\tabcolsep) * \real{0.1127}}@{}}
\toprule\noalign{}
\begin{minipage}[b]{\linewidth}\centering
\emph{Dataset}
\end{minipage} & \begin{minipage}[b]{\linewidth}\centering
\emph{Score type}
\end{minipage} & \begin{minipage}[b]{\linewidth}\centering
\emph{Group type}
\end{minipage} & \begin{minipage}[b]{\linewidth}\centering
$\overline{\mathbf{|\Delta Mean\ frequency|}}$

\emph{(SD)}
\end{minipage} & \begin{minipage}[b]{\linewidth}\centering
\emph{p}
\end{minipage} & \begin{minipage}[b]{\linewidth}\centering
\emph{Median}\(\mathbf{\Delta}\)
\end{minipage} & \begin{minipage}[b]{\linewidth}\centering
$\overline{\mathbf{|\Delta PLE|}}$

\emph{(SD)}
\end{minipage} & \begin{minipage}[b]{\linewidth}\centering
\emph{p}
\end{minipage} & \begin{minipage}[b]{\linewidth}\centering
\emph{Median}\(\mathbf{\Delta}\)
\end{minipage} \\
\midrule\noalign{}
\endhead
\bottomrule\noalign{}
\endlastfoot
\multirow{8}{=}{\centering\arraybackslash H-H} & \multirow{4}{=}{\centering\arraybackslash WD} & Orig. vs. ST & 0.092 (0.052) & \ensuremath{<}0.001 & 0.088 & 0.099 (0.055) & \ensuremath{<}0.001 & 0.09 \\
& & Orig. vs. SWO & 0.123 (0.059) & \ensuremath{<}0.001 & 0.124 & 0.134 (0.067) & \ensuremath{<}0.001 & 0.14 \\
& & Orig. vs. SWTP & 0.039 (0.025) & 0.002 & 0.035 & 0.066 (0.043) & 0.002 & 0.059 \\
& & Orig. vs. TR & 0.473 (0.080) & \ensuremath{<}0.001 & 0.479 & 0.496 (0.104) & \ensuremath{<}0.001 & 0.506 \\
& \multirow{4}{=}{\centering\arraybackslash SBERT} & Orig. vs. ST & 0.011 (0.008) & 0.008 & 0.003 & 0.086 (0.065) & 0.012 & 0.028 \\
& & Orig. vs. SWO & 0.013 (0.010) & 0.006 & -0.007 & 0.123 (0.093) & 0.008 & -0.092 \\
& & Orig. vs. SWTP & 0.014 (0.011) & 0.008 & -0.005 & 0.129 (0.101) & 0.008 & -0.054 \\
& & Orig. vs. TR & 0.034 (0.013) & \ensuremath{<}0.001 & -0.034 & 0.208 (0.116) & \ensuremath{<}0.001 & -0.171 \\
\multirow{8}{=}{\centering\arraybackslash H-TTS} & \multirow{4}{=}{\centering\arraybackslash WD} & Orig. vs. ST & 0.033 (0.033) & 0.004 & -0.003 & 0.063 (0.043) & 0.003 & 0.055 \\
& & Orig. vs. SWO & 0.060 (0.048) & 0.002 & 0.054 & 0.097 (0.065) & 0.001 & 0.098 \\
& & Orig. vs. SWTP & 0.052 (0.032) & 0.001 & 0.049 & 0.116 (0.062) & 0.001 & 0.113 \\
& & Orig. vs. TR & 0.543 (0.036) & \ensuremath{<}0.001 & 0.537 & 0.602 (0.060) & \ensuremath{<}0.001 & 0.595 \\
& \multirow{4}{=}{\centering\arraybackslash SBERT} & Orig. vs. ST & 0.003 (0.002) & 0.01 & -0.001 & 0.326 (0.241) & 0.01 & -0.02 \\
& & Orig. vs. SWO & 0.003 (0.003) & 0.01 & 0.001 & 0.397 (0.318) & 0.011 & 0.049 \\
& & Orig. vs. SWTP & 0.004 (0.003) & 0.008 & 0.002 & 0.427 (0.326) & 0.009 & 0.017 \\
& & Orig. vs. TR & 0.003 (0.003) & 0.002 & -0.002 & 0.497 (0.330) & 0.001 & -0.451 \\
\multirow{8}{=}{\centering\arraybackslash LLM-TTS} & \multirow{4}{=}{\centering\arraybackslash WD} & Orig. vs. ST & 0.036 (0.022) & 0.001 & 0.035 & 0.113 (0.064) & 0.001 & 0.116 \\
& & Orig. vs. SWO & 0.098 (0.045) & \ensuremath{<}0.001 & 0.097 & 0.110 (0.054) & \ensuremath{<}0.001 & 0.107 \\
& & Orig. vs. SWTP & 0.113 (0.042) & \ensuremath{<}0.001 & 0.111 & 0.214 (0.093) & \ensuremath{<}0.001 & 0.236 \\
& & Orig. vs. TR & 0.532 (0.041) & \ensuremath{<}0.001 & 0.531 & 0.661 (0.053) & \ensuremath{<}0.001 & 0.659 \\
& \multirow{4}{=}{\centering\arraybackslash SBERT} & Orig. vs. ST & 0.003 (0.003) & 0.002 & -0.002 & 0.516 (0.379) & 0.002 & -0.213 \\
& & Orig. vs. SWO & 0.061(0.037) & 0.01 & \ensuremath{<}0.001 & 0.500 (0.460) & 0.009 & 0.098 \\
& & Orig. vs. SWTP & 0.06(0.035) & 0.004 & \ensuremath{<}0.001 & 0.531 (0.448) & 0.005 & -0.072 \\
& & Orig. vs. TR & 0.086(0.0353) & 0.005 & -0.002 & 0.694 (0.712) & 0.003 & -0.346 \\
\end{longtable}

{\footnotesize\itshape Abbreviations: \emph{P}, combined p-value of Empiric Brown's test; \emph{Median}\(\ \Delta\): median change after shuffling; \emph{SD}, standard deviation; \emph{Orig.}, original; \emph{ST}, shuffled time; \emph{SWO}, shuffled word order; \emph{SWTP}, shuffled word-time pair; \emph{TR}, temporally randomized.\par}

\subsection{Relationship between Temporal Structure (ACW-0) and Meaning (Word Depth, generic vs specific)}

To investigate whether dynamic fluctuations carry semantically meaningful information, we analyzed the relationship between the ACW-0 and average word depths. The analysis was based on text segments of 100s with 10s intervals for all subjects (see Methods 1.8 for details).

Using the Linear Mixed Effect Model (LMM), the original texts in the H-H dataset revealed a negative relationship between ACW-0 categories and average word depths (Fixed Effect: -0.146, CI: [-0.241, -0.050], SE: 0.049, p=.003, Figure 4a). On the contrary, the shuffled texts all showed significant departures from the Original' s negative relationship: the ST, SWO, and TR conditions reversed direction to significant positive fixed effects (β = +0.058, +0.066, and +0.024 respectively; all p \ensuremath{<} .001), whereas the SWTP condition collapsed to a near-zero fixed effect with reversed direction (β = +0.007, 95\% CI [\ensuremath{<}0.001, 0.014], p = .041). (Table 4)

Such differentiation between original and shuffled texts was also observed in the H-TTS and LLM-TTS datasets. In the H-TTS dataset, only the original texts (Fixed Effect: -0.199, CI: [-0.305, -0.093], SE: 0.054, p= \ensuremath{<}.001) showed a significant negative fixed effect while shuffled time texts (Fixed Effect: -0.039, CI: [-0.045, -0.033], SE: 0.003, p= \ensuremath{<}.001, Figure 4a) displayed a weak negative relationship and others positive. (Table 4) Similarly, in the LLM-TTS dataset, the original texts exhibited a significant negative relationship (Fixed Effect: -0.168, CI: [-0.306, -0.030], SE: 0.054, p\ensuremath{<}.001). In contrast, the shuffled texts did not show significant effects. (Figure 4 and Table 4)

Together, these results mark a distinguished relationship of ACW-0 fluctuations with word depth in the original texts. Longer time windows on average include more generic words (lower word depths) while shorter time windows contain more specific words, thereby resulting in a negative correlation. This relationship is disrupted by every shuffled control, either reversing to a positive fixed effect (ST, SWO, TR in H-H; SWO, SWTP, TR in H-TTS), collapsing to near zero (SWTP in H-H), or becoming non-significant (all shuffles in LLM-TTS).

Given the presence of negative relationships across different datasets, the fixed effects of ACW-50, ACW-40, ACW-30, ACW-20, and ACW-10 were further analysed to evaluate detailed differences as supplement to ACW-0. The three datasets differ in their trend of fixed effects along the ACW variables. (Figure 4b and supplement result 5) These results suggest that, despite the analogous negative relationship of ACW-0 in the original texts of all datasets, they nevertheless seem to vary in their progression of their autocorrelation functions, including ACW-50 etc.

\begin{figure}[htbp]
\centering
\includegraphics[width=\linewidth]{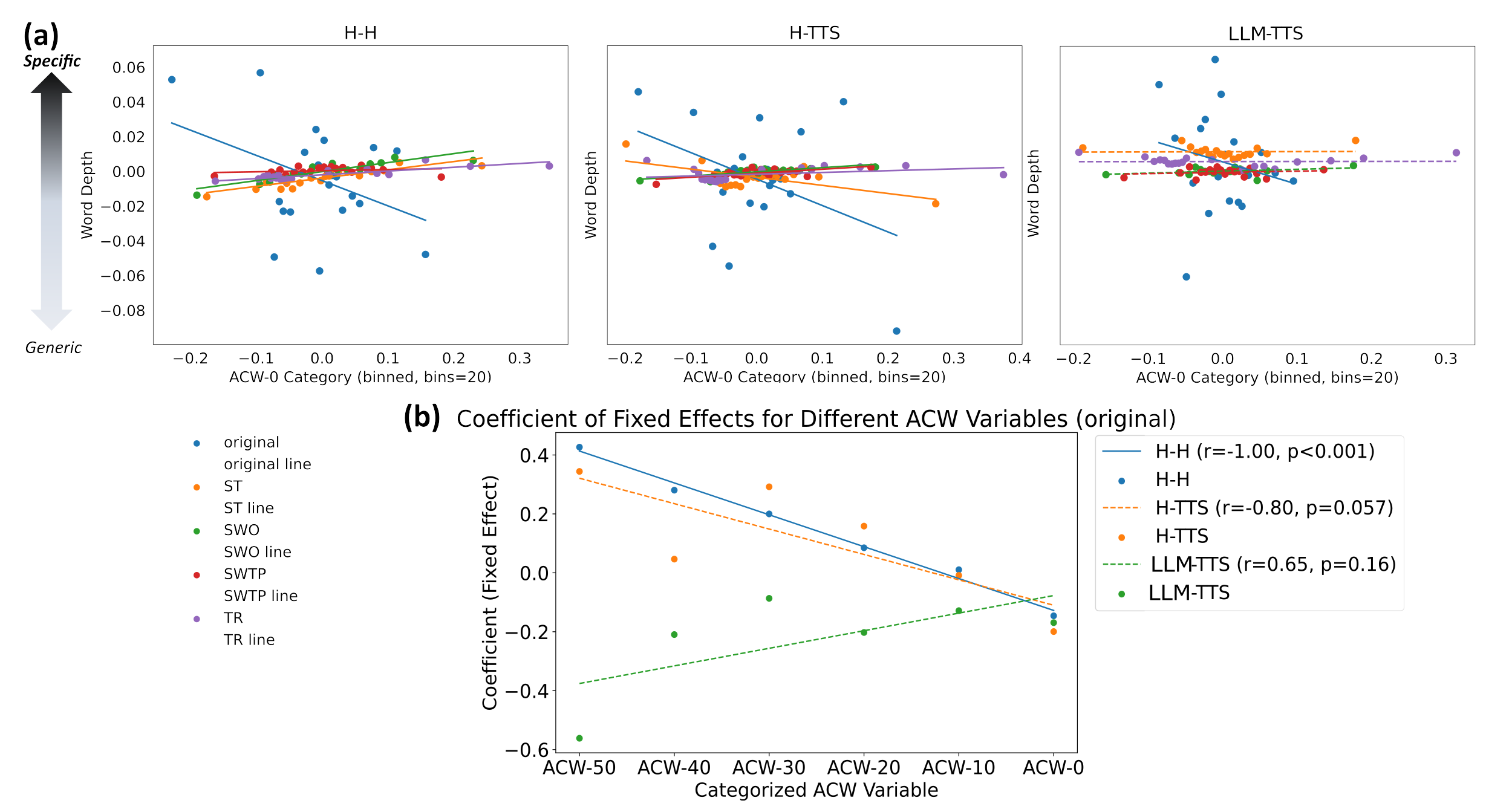}
\caption{LMM results. Due to the multitude of datapoints, the ACW-0 categories were binned (bins=20) for optimal visualization. The average word depth of each bin was calculated and plotted. The LMM results were not based on the binned ACW-0 categories. As shown in all three datasets, only the original texts show negative relationship between the two: longer ACW-0s are related to lower word depths, namely, more generic words. Continuous lines represent significant correlations (p-value \ensuremath{<} .05), whereas dashed lines represent non-significant correlations (p \ensuremath{\ge} .05) b. To further evaluate if H-H show differences with other ACW variables. The fixed effects between various ACW variables and word depths are plotted. As, shown, the decreasing trend of fixed effects is significant only in H-H dataset}
\label{fig:lmm}
\end{figure}

{\centering\small\textbf{Table 4.} The Linear Mixed Effect results between ACW-0 categories and word depths for three different datasets.\par}\nobreak\smallskip
\begin{longtable}[]{@{}
  >{\centering\arraybackslash}p{(\linewidth - 14\tabcolsep) * \real{0.1071}}
  >{\centering\arraybackslash}p{(\linewidth - 14\tabcolsep) * \real{0.1112}}
  >{\centering\arraybackslash}p{(\linewidth - 14\tabcolsep) * \real{0.0878}}
  >{\centering\arraybackslash}p{(\linewidth - 14\tabcolsep) * \real{0.1755}}
  >{\centering\arraybackslash}p{(\linewidth - 14\tabcolsep) * \real{0.0839}}
  >{\centering\arraybackslash}p{(\linewidth - 14\tabcolsep) * \real{0.1596}}
  >{\centering\arraybackslash}p{(\linewidth - 14\tabcolsep) * \real{0.0915}}
  >{\centering\arraybackslash}p{(\linewidth - 14\tabcolsep) * \real{0.1451}}@{}}
\toprule\noalign{}
\begin{minipage}[b]{\linewidth}\centering
\textbf{Dataset}
\end{minipage} & \begin{minipage}[b]{\linewidth}\centering
\textbf{Text Type}
\end{minipage} & \begin{minipage}[b]{\linewidth}\centering
\textbf{DF}
\end{minipage} & \begin{minipage}[b]{\linewidth}\centering
\textbf{Fixed Effect (β1)}
\end{minipage} & \begin{minipage}[b]{\linewidth}\centering
\textbf{SE}
\end{minipage} & \begin{minipage}[b]{\linewidth}\centering
\textbf{95\% CI}
\end{minipage} & \begin{minipage}[b]{\linewidth}\centering
\textbf{P-value}
\end{minipage} & \begin{minipage}[b]{\linewidth}\centering
\textbf{R\textsuperscript{2}conditional}
\end{minipage} \\
\midrule\noalign{}
\endfirsthead
\toprule\noalign{}
\begin{minipage}[b]{\linewidth}\centering
\textbf{Dataset}
\end{minipage} & \begin{minipage}[b]{\linewidth}\centering
\textbf{Text Type}
\end{minipage} & \begin{minipage}[b]{\linewidth}\centering
\textbf{DF}
\end{minipage} & \begin{minipage}[b]{\linewidth}\centering
\textbf{Fixed Effect (β1)}
\end{minipage} & \begin{minipage}[b]{\linewidth}\centering
\textbf{SE}
\end{minipage} & \begin{minipage}[b]{\linewidth}\centering
\textbf{95\% CI}
\end{minipage} & \begin{minipage}[b]{\linewidth}\centering
\textbf{P-value}
\end{minipage} & \begin{minipage}[b]{\linewidth}\centering
\textbf{R\textsuperscript{2}conditional}
\end{minipage} \\
\midrule\noalign{}
\endhead
\bottomrule\noalign{}
\endlastfoot
\multirow{5}{=}{\centering\arraybackslash H-H} & Orig. & 412 & -0.146 & 0.049 & [-0.241, -0.050] & 0.003 & 0.046 \\
& ST & 110594 & 0.058 & 0.003 & [0.052, 0.063] & \ensuremath{<}0.001 & 0.025 \\
& SWO & 103649 & 0.066 & 0.003 & [0.060, 0.072] & \ensuremath{<}0.001 & 0.029 \\
& SWTP & 86462 & 0.007 & 0.003 & [\ensuremath{<}0.001, 0.014] & 0.041 & 0.034 \\
& TR & 380539 & 0.024 & 0.002 & [0.021, 0.027] & \ensuremath{<}0.001 & 0.003 \\
\multirow{5}{=}{\centering\arraybackslash H-TTS} & Orig. & 329 & -0.199 & 0.054 & [-0.305, -0.093] & \ensuremath{<}0.001 & 0.065 \\
& ST & 76081 & -0.039 & 0.003 & [-0.045, -0.033] & \ensuremath{<}0.001 & 0.051 \\
& SWO & 27175 & 0.021 & 0.003 & [0.015, 0.027] & \ensuremath{<}0.001 & 0.057 \\
& SWTP & 11148 & 0.012 & 0.003 & [0.005, 0.018] & \ensuremath{<}0.001 & 0.067 \\
& TR & 368713 & 0.014 & 0.002 & [0.010, 0.017] & \ensuremath{<}0.001 & 0.003 \\
\multirow{5}{=}{\centering\arraybackslash LLM-TTS} & Orig. & 195 & -0.168 & 0.070 & [-0.306, -0.030] & 0.002 & 0.318 \\
& ST & 41602 & \ensuremath{<}0.001 & \ensuremath{<}0.001 & [-0.001, 0.001] & 0.636 & 0.306 \\
& SWO & 5072 & \ensuremath{<}0.001 & 0.001 & [-0.002, 0.002] & 0.986 & 0.204 \\
& SWTP & 5099 & \ensuremath{<}0.001 & 0.001 & [-0.002, 0.002] & 0.852 & 0.223 \\
& TR & 228173 & 0.003 & 0.002 & [-0.001, 0.007] & 0.179 & 0.019 \\
\end{longtable}

In addition, to test whether the results were driven by uneven WordNet hierarchies, particularly the relatively shallow verb hierarchy, we reanalyzed the H-H data after removing all verbs. The original H-H condition remained significantly negative (coef = -0.1867, \emph{p} = 0.000242), whereas the main sign pattern across disrupted controls was preserved. Recomputing the analysis using a newer English WordNet (OEWN 2025; McCrae et al., 2025) likewise yielded the same qualitative result, with the original condition again remaining negative and significant (coef = -0.1123, \emph{p} = 0.0173). We additionally tested an equal-duration control in which every word was assigned a uniform duration of 0.5s while preserving word order and lexical content; under this control, the negative relationship was abolished (coef = -0.0259, p = 0.438), converging with the TR condition in indicating that natural word-timing variation is integral to the observed effect. Finally, although WordNet depth showed a weak negative association with Zipf frequency (\emph{r} = -0.076, \emph{p} = 0.002), the ACW effect remained significant and of similar magnitude after adjustment for mean Zipf frequency. Full results are reported in Supplementary Results 6--8.

\section{Discussion}

Are there systematic, meaningful timescales in the semantics of speech that can be measured with autocorrelation windows? We addressed this question by (i) constructing high-resolution semantic time series from spoken language and (ii) quantifying their temporal structure using ACW-0 and complementary spectral metrics. Across human narratives, conventional text-to-speech (TTS), and LLM-derived TTS, we found that semantic fluctuations are not temporally random: original signals differ reliably from shuffled controls in both time and frequency domains, and these differences cannot be reproduced by shuffling alone. Moreover, we observed a robust coupling between the duration of semantic timescales (ACW-0) and lexical specificity, indexed by WordNet depth, which disappears or reverses once temporal structure is destroyed. Taken together, these results suggest that semantic content in natural speech is organized along intrinsic timescales that depend on how generic or specific the language is.

\subsection*{Main finding: Semantic fluctuations exhibit intrinsic timescales}

A growing body of information-theoretic work has characterized how information is distributed across linguistic units --- across ordinal positions of tokens, sentences, or paragraphs in text (Genzel \& Charniak, 2002; Jaeger \& Levy, 2007; Giulianelli et al., 2021; Tsipidi et al., 2024, 2025), and across the duration of individual speech units as a function of local predictability (Aylett \& Turk, 2004, 2006; Bell et al., 2003). These studies establish that the distribution of information in language is not random but systematically structured. What has received less attention is whether this temporal organization also manifests at the level of the continuous semantic signal unfolding in wall-clock spoken time, characterized directly with time-series tools instead of token position or per-word duration. Our first main finding is that semantic time series derived from word depth and SBERT similarity show robust slow--fast structure, distinguishable from a range of shuffled controls. Using dynamic time warping, ACW-0, mean spectral frequency, and power-law exponents, we observed consistent differences between original and shuffled signals across all three datasets. Permutation tests further showed that these differences do not arise among shuffled signals themselves, indicating that both the ordering of time points and the arrangement of words contribute to the observed timescales. In other words, semantic signals extracted from speech carry temporal memory beyond what would be expected from their marginal distributions or average word durations.

\subsection*{Disentangling Specificity from Timescales: Addressing Potential Circularity}

A natural concern is circularity: because both our predictor (lexical specificity) and our timescale index are derived from the same semantic stream, might we simply be correlating a measure with itself? Our analyses argue against this interpretation. The mean WordNet depth per window is a level measure---a static summary of how generic or specific the vocabulary is on average---and is independent of word order and timing. By contrast, ACW-0 is a dynamic measure that captures how long fluctuations in specificity persist before decorrelating. Two windows can therefore have identical average depth but very different ACW-0 values, depending on how generic and specific items are sequenced in time.

Empirically, we found a robust negative relationship between ACW-0 and average word depth in the original signals: windows with longer ACW-0 are enriched in generic, low-depth words, whereas windows with shorter ACW-0 contain more specific, high-depth words. Crucially, this relationship is disrupted or reversed under shuffled controls that preserve the global distribution of word depths but alter lexical order and timing. This pattern would not be expected if our effect merely reflected a trivial property of the depth values themselves (e.g., mean or variance). Instead, it points to a coupling between what is said (generic vs. specific) and how it fluctuates over time (long vs. short semantic persistence), supporting the view that level and dynamics are separable yet systematically related dimensions of semantic organization.

\subsection*{Semantic timescales, information content, and lossy contextual memory}

Prior information-theoretic work typically defines the information content of a linguistic unit in terms of its surprisal, i.e., the negative log probability of the current word given its preceding context, \(\text{Surprisal}(w_{t}) = - \log P(w_{t} \mid c_{\left. <t \right.\ })\)(Hale, 2001; Jaeger, 2010). A related lexical-semantic tradition defines information content (IC) in formally similar terms, \(\text{IC}(c) = - \log p(c)\), such that lower-probability concepts carry greater IC; in WordNet-like taxonomic systems, this notion is closely tied to specificity, because more specific concepts tend to be lower-probability and thus higher-IC than more generic ones (Resnik, 1995; Seco et al., 2004).

Within this broader framework, our findings suggest that semantic specificity is not only a lexical property but is also organized across time. Importantly, ACW-0 is not a measure of word duration, nor does it merely indicate that specific words occupy longer or shorter stretches of acoustic time. Rather, ACW-0 indexes the persistence of a dynamic semantic pattern: shorter ACW-0 reflects faster updating and weaker persistence of contextual semantic structure, whereas longer ACW-0 reflects more stable contextual support.

Read in this way, our results align naturally with lossy-context surprisal approaches to language processing, in which prediction depends on the recoverability of prior context representation (Futrell et al., 2020; Hahn et al., 2021, 2022). Surprisal is formalized as \(D_{lc\text{-}surprisal}(w_{i} \mid r_{i - 1}) \propto - \log p(w_{i} \mid r_{i - 1})\), where \(r_{i - 1}\)is a lossy memory representation of prior context. On that view, reduced contextual recoverability weakens the effective context and elevates the surprisal of upcoming material; related work formalizes this as a broader memory--surprisal trade-off, in which lower surprisal requires retaining more predictive information from the recent past (Hahn et al., 2021, 2022). We treat ACW-0 as a complementary, signal-level index of this temporal organization --- a measure of how persistently the semantic state remains predictable from its own recent past. Therefore, when a segment has lower ACW-0, hence shorter predictable memory, it would be natural for the segment to contain more words with higher surprisal, namely, more specific words in the taxonomy context with higher information content.

This interpretation is also compatible with production-side evidence that speakers are probability-sensitive during incremental production, using less informative expressions for more predictable meanings and more informative expressions --- including more specific referring expressions --- when events or referents are less predictable (Levy \& Jaeger, 2007; Jaeger, 2010; Tily \& Piantadosi, 2009; Stegemann-Philipps et al., 2021; Futrell, 2023). Under this view, the association we observe between shorter semantic timescales and more specific lexical content fits a broader context-sensitive perspective in which semantic specificity, information content, and prediction are jointly constrained by the persistence of contextual memory over time. We note that discourse organization itself is known to systematically shape both surprisal contours (Tsipidi et al., 2024, 2025) and lexical/topical structure (Hearst, 1997; Morris \& Hirst, 1991), so the coupling between ACW-0 and mean depth may also reflect a shared dependence on higher-level discourse structure; disentangling these pathways is a natural direction for future work.

\subsection*{Situating semantic timescales within neural accounts of language processing}

The observation of timescales in language is well in line with other data. Scale-free dynamics and multiple interacting timescales in language have been reported at acoustic and behavioral levels (Kello, 2004; Kello et al., 2010; Dale et al., 2016; Torre et al., 2019; Corral \& Serra, 2020; Hasson et al., 2018; Honey et al., 2016). Our findings extend this work to semantic content itself, using both WordNet-based word depth and SBERT-based sentence vectors. We show that segments with more persistent semantic fluctuations (longer ACW-0) are enriched in generic, low-depth words, whereas segments with rapidly decorrelating fluctuations (shorter ACW-0) preferentially contain specific, high-depth words. While our design does not allow us to claim that timescales cause these semantic properties, the comparison with shuffled controls indicates that their coupling is not a trivial by-product of word duration or vocabulary composition.

Current neural accounts of higher-order semantics help locate where our contribution sits. Decoding frameworks map distributed brain activity onto semantic dimensions, showing that the representational format recovered depends on both brain region and analytic approach (Frisby et al., 2023), while neurocomputational models such as ROSE propose candidate neural operations through which hierarchical linguistic structure is assembled in real time (Murphy, 2025). Our findings add a speech-level constraint to this picture: the temporal deployment of generic versus specific content in the output signal is not random but follows measurable timescale structure. This structure could serve as a target observable for decoding models---testing whether brain regions with longer intrinsic timescales preferentially encode the sustained, generic semantic content we identify---and as an output-side benchmark for neurocomputational models that predict how hierarchical language processing should shape the temporal statistics of speech. Recent work by Kumar et al. (2025), which applied a closely related framework to fMRI data and found that information transfer from speech semantics to intrinsic neuronal timescales emerges specifically at the level of timescales, offers an encouraging first step in connecting the speech-level regularities we describe to their neural counterparts.

Taken together, language use is closely tied to different temporal integration windows: generic/topic-level material is preferentially deployed across longer windows, whereas specific, detail-rich content tends to be clustered within shorter windows. The present results demonstrate that this organization is already present in the speech signal---a prerequisite for any account in which the brain exploits input timescales during comprehension. This is compatible with the broader proposal that input, brain activity, and behavior share partially aligned timescales (Northoff et al., 2020; Ao et al., 2025), though our data speak only to the input side of that framework. Whether the specificity--timescale coupling we observe reflects hierarchical production mechanisms or is exploited by neural timescale hierarchies during comprehension remains to be tested.

\subsection*{Limitations of the study}

While the ideal scenario would involve a perfect conversion from audio to text with accurate timestamps, variations in pronunciation and the inherent limitations of the speech-to-text model may introduce deviations. Future research could focus on fine-tuning these models to improve timestamp accuracy.

Our predictor indexes WordNet-based lexical specificity and is interpreted ordinally because WordNet's structure is uneven in depth and density (Wang \& Hirst, 2011). In particular, branch depth and density vary across the taxonomy, verb hierarchies are comparatively shallow, and abstract regions are less consistently stratified; such properties can inflate variance in absolute depth. Future work could incorporate context-sensitive word sense disambiguation, although our sensitivity analysis using the Lesk algorithm yielded results consistent with the sense-averaged measure used here. We therefore restrict interpretation to relative contrasts along the generic--specific axis. Additionally, as silences in the audio do not possess word depths, short intervals of silence were excluded from the word depth fluctuations. An alternative approach could be assigning values such as the average word depth or zeros to these intervals, potentially smoothing the fluctuations.

Our SBERT analysis uses one reasonable choice of embedding model and window parameters but is not the only and absolute optimal one; systematic exploration of alternative embedding models and window settings is a natural direction for future work.

Finally, to enhance the generalizability of these findings, future studies should consider including a larger sample of subjects and a broader range of speech types.

\section{Conclusion}

In this study, we investigated whether semantic content in spoken language exhibits non-random temporal organization when represented as wall-clock-aligned semantic time series. We developed and applied a semantic-timescale pipeline combining WordNet-based lexical specificity, SBERT-based contextual similarity, autocorrelation-window measures, spectral descriptors, and multiple shuffled controls. Across human-read narratives, human-derived TTS, and LLM-derived TTS, both WordNet-depth and SBERT-similarity signals showed temporal structure that differed from disrupted controls.

We further found that the ACW-0 of the WordNet-depth signal was systematically related to lexical specificity: longer semantic timescales were associated with more generic, lower-depth lexical material, whereas shorter semantic timescales were associated with more specific, higher-depth lexical material. This relationship was attenuated, abolished, or reversed when temporal order, word timing, or word--duration structure was disrupted, suggesting that the effect is not reducible to static lexical distributions alone.

We interpret these findings cautiously. WordNet depth and SBERT similarity are coarse semantic proxies, and the LLM-derived TTS condition does not provide a clean test of human versus AI language representations because timing is jointly shaped by generated text and speech synthesis. Nevertheless, the results support the usefulness of semantic-timescale measures as interpretable descriptors of how semantic content is deployed over spoken time. Future work should test alternative semantic signals, larger and more diverse speech corpora, and direct links between speech-level semantic timescales and neural models of language processing.

\section*{Declaration of Competing Interest}

The authors report no conflicts of interest, financial or personal, that may have influenced the conduct or findings of this study.

\section*{Acknowledgements}

The authors would like to thank Chun-Hung Chen for providing helpful conceptual ideas during the development of this work.

\section*{Data availability}

Data will be made available on request.

\section*{References}

\begin{enumerate}[label=\arabic*.,leftmargin=2.4em,labelsep=0.6em,itemsep=2pt,topsep=4pt,align=left]
\item
  Çatal, Y., Keskin, K., Wolman, A., Klar, P., Smith, D., \& Northoff, G. (2024). Flexibility of intrinsic neural timescales during distinct behavioral states. \emph{Communications Biology, 7}(1), 1--17. \url{https://doi.org/10.1038/s42003-024-07349-1}
\item
  Henzler-Wildman, K. A., Lei, M., Thai, V., Kerns, S. J., Karplus, M., \& Kern, D. (2007). A hierarchy of timescales in protein dynamics is linked to enzyme catalysis. \emph{Nature, 450}(7171), 913--916. \url{https://doi.org/10.1038/nature06407}
\item
  Regev, T.I., Casto, C., Hosseini, E.A. et al. (2024). Neural populations in the language network differ in the size of their temporal receptive windows. \emph{Nat Hum Behav}. \url{https://doi.org/10.1038/s41562-024-01944-2}
\item
  Scheffer, M., \& Carpenter, S. R. (2003). Catastrophic regime shifts in ecosystems: Linking theory to observation. \emph{Trends in Ecology \& Evolution, 18}(12), 648--656. \url{https://doi.org/10.1016/j.tree.2003.09.002}
\item
  Wolff A, Berberian N, Golesorkhi M, Gomez-Pilar J, Zilio F, Northoff G. (2022). Intrinsic neural timescales: temporal integration and segregation. \emph{Trends Cogn Sci, 26}(2), 159-173. doi: 10.1016/j.tics.2021.11.007.
\item
  Hasson, U., Yang, E., Vallines, I., Heeger, D. J., \& Rubin, N. (2008). A hierarchy of temporal receptive windows in human cortex. \emph{The Journal of neuroscience : the official journal of the Society for Neuroscience, 28}(10), 2539--2550. \url{https://doi.org/10.1523/JNEUROSCI.5487-07.2008}
\item
  Lerner, Y., Honey, C. J., Silbert, L. J., \& Hasson, U. (2011). Topographic mapping of a hierarchy of temporal receptive windows using a narrated story. \emph{Journal of Neuroscience, 31}(8), 2906-2915. \url{https://doi.org/10.1523/JNEUROSCI.3684-10.2011}.
\item
  Himberger, K. D., Chien, H.-Y., \& Honey, C. J. (2018). Principles of Temporal Processing Across the Cortical Hierarchy. \emph{Neuroscience, 389}, 161--174. \url{https://doi.org/10.1016/j.neuroscience.2018.04.030}
\item
  Honey, C. J., Thesen, T., Donner, T. H., Silbert, L. J., Carlson, C. E., Devinsky, O., Doyle, W. K., Rubin, N., Heeger, D. J., \& Hasson, U. (2012). Slow cortical dynamics and the accumulation of information over long timescales. \emph{Neuron, 76}(2), 423--434. \url{https://doi.org/10.1016/j.neuron.2012.08.011}
\item
  Levelt, W. J. M. (1989). \emph{Speaking: From intention to articulation}. MIT Press.
\item
  Levelt, W. J. M., Roelofs, A., \& Meyer, A. S. (1999). A theory of lexical access in speech production. \emph{Behavioral and Brain Sciences, 22}(1), 1--38. \url{https://doi.org/10.1017/S0140525X99001776}
\item
  Levelt, W.J.M. (2001). Spoken word production: A theory of lexical access. \emph{Proc. Natl. Acad. Sci. U.S.A. 98}(23), 13464-13471. \url{https://doi.org/10.1073/pnas.231459498}
\item
  Verwoert, M., Amigó-Vega, J., Gao, Y. et al. (2025). Whole-brain dynamics of articulatory, acoustic and semantic speech representations. \emph{Commun Biol, 8}, 432. \url{https://doi.org/10.1038/s42003-025-07862-x}
\item
  Slone, L. K., Abney, D. H., Smith, L. B., \& Yu, C. (2023). The temporal structure of parent talk to toddlers about objects. \emph{Cognition, 230}, 105266. \url{https://doi.org/10.1016/j.cognition.2022.105266}
\item
  Sandler, M., Choung, H., Ross, A., \& David, P. (2025). A Linguistic Comparison Between Human and ChatGPT-Generated Conversations. In C. Wallraven, C.-L. Liu, \& A. Ross (Eds.), \emph{Pattern Recognition and Artificial Intelligence} (pp. 366--380). Springer Nature. \url{https://doi.org/10.1007/978-981-97-8702-9_25}
\item
  Ferrer-i-Cancho, R. (2005). The variation of Zipf's law in human language. \emph{European Physical Journal B, 44}, 249--257. \url{https://doi.org/10.1140/epjb/e2005-00121-8}
\item
  Neophytou, K., van-Egmond, M., \& Avrutin, S. (2017). Zipf's law in aphasia across languages: A comparison of English, Hungarian, and Greek. \emph{Journal of Quantitative Linguistics, 24}, 178--196. \url{https://doi.org/10.1080/09296174.2016.1263786}
\item
  Zipf, G. K. (1935). \emph{The Psycho-Biology of Language}. Houghton-Mifflin.
\item
  Zipf, G. K. (1949). \emph{Human Behavior and the Principle of Least Effort}. Addison-Wesley.
\item
  Kuraszkiewicz, W., \& Łukaszewicz, J. (1951). Ilość różnych wyrazów w zależności od długości tekstu. \emph{Pamiętnik Literacki}.
\item
  Heaps, H. S. (1978). \emph{Information retrieval: Computational and theoretical aspects}. Academic Press.
\item
  Wang, Yanru and Chen, Xinying. "Structural Complexity of Simplified Chinese Characters". \emph{Recent Contributions to Quantitative Linguistics}, edited by Arjuna Tuzzi, Martina Benesová and Ján Macutek, Berlin, München, Boston: De Gruyter Mouton, 2015, pp. 229-240. \url{https://doi.org/10.1515/9783110420296-019}
\item
  Sanada, H. (2011). Investigations in Japanese Historical Lexicology. \emph{Japanese Language \& Literature, 45}(2), 519.
\item
  Zipf, G. K. (1932). \emph{Selected Studies of the Principle of Relative Frequencies of Language}. Cambridge, Massachusetts: Harvard University Press.
\item
  Boroda, M., \& Altmann, G. (1991). A First Glance at Textual Statistics. \emph{Glottometrika, 12}.
\item
  Grégoire, A. (1899). Variation de la durée de la syllabe française suivant sa place dans les groupements phonétiques. \emph{La Parole}.
\item
  Grzybek, P., Stadlober, E., \& Kelih-Emmerich, N. (2007). The relationship of word length and sentence length: The inter-textual perspective. In \emph{Advances in data analysis} (pp. 611-618). Springer.
\item
  Mačutek, J., Chromý, J., \& Koščová, M. (2018). Menzerath-Altmann Law and Prothetic /v/ in Spoken Czech.~\emph{Journal of Quantitative Linguistics},~\emph{26}(1), 66--80. \url{https://doi.org/10.1080/09296174.2018.1424493}
\item
  Fletcher, J. (2010). The Prosody of Speech: Timing and Rhythm. In \emph{The Handbook of Phonetic Sciences} (eds W.J. Hardcastle, J. Laver and F.E. Gibbon). \url{https://doi.org/10.1002/9781444317251.ch15}
\item
  Blum, F., Paschen, L., Forkel, R. et al. (2024). Consonant lengthening marks the beginning of words across a diverse sample of languages. \emph{Nat Hum Behav, 8}, 2127--2138. \url{https://doi.org/10.1038/s41562-024-01988-4}
\item
  Holmlund, T. B., Chandler, C., Foltz, P. W., Diaz-Asper, C., Cohen, A. S., Rodriguez, Z., \& Elvevåg, B. (2023). Towards a temporospatial framework for measurements of disorganization in speech using semantic vectors. \emph{Schizophrenia Research, 259}, 71--79. \url{https://doi.org/10.1016/j.schres.2022.09.020}
\item
  Palominos, C., He, R., Fröhlich, K., Mülfarth, R. R., Seuffert, S., Sommer, I. E., Homan, P., Kircher, T., Stein, F., \& Hinzen, W. (2024). Approximating the semantic space: Word embedding techniques in psychiatric speech analysis. \emph{Schizophrenia, 10}(1), 1--10. \url{https://doi.org/10.1038/s41537-024-00524-7}
\item
  Miller, G. A. (1995). WordNet: A lexical database for English. \emph{Communications of the ACM, 38}(11), 39--41. doi:10.1145/219717.219748.
\item
  Princeton University. (2010). \emph{WordNet}. Retrieved from \url{https://wordnet.princeton.edu/}.
\item
  Reimers, N., \& Gurevych, I. (2019). Sentence-BERT: Sentence embeddings using Siamese BERT-networks. \emph{arXiv}. \url{https://doi.org/10.48550/arXiv.1908.10084}
\item
  Honey, C. J., Chen, J., Müsch, K., \& Hasson, U. (2016). How long is now? The multiple timescales of language processing. \emph{Behavioral and Brain Sciences, 39}, e77. \url{https://doi.org/10.1017/S0140525X15000825}.
\item
  Golesorkhi, M., Gomez-Pilar, J., Tumati, S. et al. (2021). Temporal hierarchy of intrinsic neural timescales converges with spatial core-periphery organization. \emph{Commun Biol, 4}, 277. \url{https://doi.org/10.1038/s42003-021-01785-z}
\item
  Lewis GA, Poeppel D, Murphy GL. (2015). The neural bases of taxonomic and thematic conceptual relations: an MEG study. \emph{Neuropsychologia, 68}, 176-89. doi: 10.1016/j.neuropsychologia.2015.01.011.
\item
  Jamali, M., Grannan, B., Cai, J. et al. (2024). Semantic encoding during language comprehension at single-cell resolution. \emph{Nature, 631}, 610--616. \url{https://doi.org/10.1038/s41586-024-07643-2}
\item
  Crossley, S., Salsbury, T., \& McNamara, D. (2009). Measuring L2 lexical growth using hypernymic relationships. \emph{Language Learning, 59}(2), 307--334. \url{https://doi.org/10.1111/j.1467-9922.2009.00508.x}
\item
  Crossley, S. A., \& McNamara, D. S. (2012). Predicting second language writing proficiency: The roles of cohesion and linguistic sophistication. \emph{Journal of Research in Reading, 35}(2), 115--135. \url{https://doi.org/10.1111/j.1467-9817.2010.01449.x}
\item
  Bolognesi, M., Burgers, C. \& Caselli, T. (2020). On abstraction: decoupling conceptual concreteness and categorical specificity. \emph{Cogn Process, 21}, 365--381. \url{https://doi.org/10.1007/s10339-020-00965-9}
\item
  Paivio, A. (1991). Dual coding theory: Retrospect and current status. \emph{Canadian Journal of Psychology / Revue canadienne de psychologie, 45}(3), 255--287. \url{https://doi.org/10.1037/h0084295}
\item
  Sadoski, M., \& Paivio, A. (1994). A dual coding view of imagery and verbal processes in reading comprehension. In R. B. Ruddell, M. R. Ruddell, \& H. Singer (Eds.), \emph{Theoretical models and processes of reading} (4th ed., pp. 582--601). International Reading Association.
\item
  Binder, J. R., Conant, L. L., Humphries, C. J., Fernandino, L., Simons, S. B., Aguilar, M., \& Desai, R. H. (2016). Toward a brain-based componential semantic representation.~\emph{Cognitive neuropsychology},~\emph{33}(3-4), 130--174. \url{https://doi.org/10.1080/02643294.2016.1147426}
\item
  Hnazaee, M. F., Khachatryan, E., Chehrazad, S., Kotarcic, A., De Letter, M., \& Van Hulle, M. M. (2020). Overlapping connectivity patterns during semantic processing of abstract and concrete words revealed with multivariate Granger Causality analysis. \emph{Scientific Reports, 10}(1), 2803. \url{https://doi.org/10.1038/s41598-020-59473-7}
\item
  Bi Y. (2021). Dual coding of knowledge in the human brain. \emph{Trends Cogn Sci, 25}(10), 883-895. doi: 10.1016/j.tics.2021.07.006.
\item
  Vignali, L., Xu, Y., Turini, J., Collignon, O., Crepaldi, D., \& Bottini, R. (2023). Spatiotemporal dynamics of abstract and concrete semantic representations. \emph{Brain and language, 243}, 105298. \url{https://doi.org/10.1016/j.bandl.2023.105298}
\item
  Müller, Meinard (2007). Dynamic Time Warping. In \emph{Information Retrieval for Music and Motion}, chapter 4, pages 69-84. Springer. doi:10.1007/978-3-540-74048-3.
\item
  García, A.M., \& Ibáñez, A. (Eds.). (2022). \emph{The Routledge Handbook of Semiosis and the Brain} (1st ed.). Routledge. \url{https://doi.org/10.4324/9781003051817}
\item
  Smith, D., Wolff, A., Wolman, A., Ignaszewski, J., \& Northoff, G. (2022). Temporal continuity of self: Long autocorrelation windows mediate self-specificity. \emph{NeuroImage, 257}, 119305. \url{https://doi.org/10.1016/j.neuroimage.2022.119305}
\item
  OpenAI. (2024). \emph{GPT-4} [Large language model]. \url{https://openai.com}
\item
  Pawar, A., \& Mago, V. (2018). Calculating the similarity between words and sentences using a lexical database and corpus statistics. \emph{International Journal of Computer Applications, 182}(34), 1--10.
\item
  Theiler, J., \& Prichard, D. (1996). Constrained-realization Monte-Carlo method for hypothesis testing. \emph{Physica D: Nonlinear Phenomena, 94}(4), 221--235. \url{https://doi.org/10.1016/0167-2789(96)00050-4}
\item
  Brown, M. B. (1975). 400: A Method for Combining Non-Independent, One-Sided Tests of Significance. \emph{Biometrics, 31}(4), 987--92. \url{https://doi.org/10.2307/2529826}
\item
  Poole, W., Gibbs, D. L., Shmulevich, I., Bernard, B., \& Knijnenburg, T. A. (2016). Combining dependent P-values with an empirical adaptation of Brown' s method. \emph{Bioinformatics (Oxford, England), 32}(17), i430--i436. \url{https://doi.org/10.1093/bioinformatics/btw438}
\item
  dtaidistance. (2023). \emph{Dynamic Time Warping (DTW) in Python}. Available at: \url{https://github.com/wannesm/dtaidistance}
\item
  MathWorks. (2022). \emph{Signal Processing Toolbox for MATLAB}. The MathWorks, Inc. Available at: \url{https://www.mathworks.com/products/signal.html}
\item
  Seabold, S. and Perktold, J. (2010). Statsmodels: Econometric and Modeling with Python. \emph{9th Python in Science Conference}, Austin, 28 June-3 July, 2010, 57-61. \url{https://doi.org/10.25080/Majora-92bf1922-011}
\item
  Murray, J., Bernacchia, A., Freedman, D. et al. (2014). A hierarchy of intrinsic timescales across primate cortex. \emph{Nat Neurosci, 17}, 1661--1663. \url{https://doi.org/10.1038/nn.3862}
\item
  Chaudhuri, R., Knoblauch, K., Gariel, M.-A., Kennedy, H., \& Wang, X.-J. (2015). A Large-Scale Circuit Mechanism for Hierarchical Dynamical Processing in the Primate Cortex. \emph{Neuron, 88}(2), 419--431. \url{https://doi.org/10.1016/j.neuron.2015.09.008}
\item
  Phipson, B., \& Smyth, G. K. (2010). Permutation P-values should never be zero: calculating exact P-values when permutations are randomly drawn. \emph{Statistical applications in genetics and molecular biology, 9}, Article39. \url{https://doi.org/10.2202/1544-6115.1585}
\item
  Kolmogorov, A. N. (1933). Sulla determinazione empirica di una legge di distribuzione. \emph{Giornale dell'Istituto Italiano degli Attuari, 4}, 83--91.
\item
  Smirnov, N. (1948). Table for Estimating the Goodness of Fit of Empirical Distributions. \emph{The Annals of Mathematical Statistics, 19}(2), 279-281. \url{https://doi.org/10.1214/aoms/1177730256}
\item
  Spearman, C. (1904). The Proof and Measurement of Association between Two Things. \emph{The American Journal of Psychology, 15}(1), 72--101. \url{https://doi.org/10.2307/1412159}
\item
  Hasson, U., Chen, J., \& Honey, C. J. (2015). Hierarchical process memory: memory as an integral component of information processing. \emph{Trends in Cognitive Sciences, 19}(6), 304-313. \url{https://doi.org/10.1016/j.tics.2015.04.006}.
\item
  Kello, C. T., Beltz, B. C., Holden, J. G., \& Van Orden, G. C. (2007). The emergent coordination of cognitive function. \emph{Journal of experimental psychology. General, 136}(4), 551--568. \url{https://doi.org/10.1037/0096-3445.136.4.551}
\item
  Janssen, N., Meij, M.v.d., López-Pérez, P.J. et al. (2020). Exploring the temporal dynamics of speech production with EEG and group ICA. \emph{Sci Rep, 10}, 3667. \url{https://doi.org/10.1038/s41598-020-60301-1}.
\item
  Chang, C. H. C., Nastase, S. A., \& Hasson, U. (2022). Information flow across the cortical timescale hierarchy during narrative construction. \emph{Proceedings of the National Academy of Sciences of the United States of America, 119}(51), e2209307119. \url{https://doi.org/10.1073/pnas.2209307119}.
\item
  Zhu, Y., Xu, M., Lu, J. et al. (2022). Distinct spatiotemporal patterns of syntactic and semantic processing in human inferior frontal gyrus. \emph{Nat Hum Behav, 6}, 1104--1111. \url{https://doi.org/10.1038/s41562-022-01334-6}.
\item
  Murphy, E., Forseth, K.J., Donos, C. et al. (2023). The spatiotemporal dynamics of semantic integration in the human brain. \emph{Nat Commun, 14}, 6336. \url{https://doi.org/10.1038/s41467-023-42087-8}
\item
  Levy, R., \& Jaeger, T. F. (2007). Speakers optimize information density through syntactic reduction. In B. Schölkopf, J. Platt, \& T. Hofmann (Eds.), \emph{Advances in Neural Information Processing Systems 19}.
\item
  Hale, J. (2001). A Probabilistic Earley Parser as a Psycholinguistic Model. In \emph{Proceedings of the Second Meeting of the North American Chapter of the Association for Computational Linguistics}.
\item
  Resnik, P. (1995). Using information content to evaluate semantic similarity in a taxonomy. In \emph{Proceedings of the 14th international joint conference on Artificial intelligence - Volume 1} (pp. 448--453). Morgan Kaufmann Publishers Inc.
\item
  Seco, Veale, \& Hayes (2004). \textbf{"An intrinsic information content metric for semantic similarity in WordNet"} in \textbf{ECAI 2004}, pp. 1089--1090
\item
  Altmann~EG, Pierrehumbert~JB, Motter~AE (2009)~Beyond Word Frequency: Bursts, Lulls, and Scaling in the Temporal Distributions of Words. PLOS ONE 4(11): e7678.~\url{https://doi.org/10.1371/journal.pone.0007678}
\item
  Kello CT, Brown GD, Ferrer-I-Cancho R, Holden JG, Linkenkaer-Hansen K, Rhodes T, Van Orden GC. (2010). Scaling laws in cognitive sciences. \emph{Trends Cogn Sci, 14}(5), 223-32. doi: 10.1016/j.tics.2010.02.005.
\item
  Dale, R., Kello, C. T., \& Schoenemann, P. T. (2016). Seeking Synthesis: The Integrative Problem in Understanding Language and Its Evolution. \emph{Topics in cognitive science, 8}(2), 371--381. \url{https://doi.org/10.1111/tops.12199}
\item
  Kello, C. T. (2004). Characterizing the evolutionary dynamics of language. \emph{Trends in Cognitive Sciences, 8}(9), 392-394.
\item
  Torre, I. G., Luque, B., Lacasa, L., Kello, C. T., \& Hernández-Fernández, A. (2019). On the physical origin of linguistic laws and lognormality in text. \emph{Royal Society Open Science, 6}(8), 191023.
\item
  Corral, Á.; Serra, I. (2020). The Brevity Law as a Scaling Law, and a Possible Origin of Zipf's Law for Word Frequencies. \emph{Entropy, 22}, 224. \url{https://doi.org/10.3390/e22020224}
\item
  Hasson, U., Egidi, G., Marelli, M., \& Willems, R. M. (2018). Grounding the neurobiology of language in first principles: The necessity of non-language-centric explanations for language comprehension. \emph{Cognition, 180}, 135-157. \url{https://doi.org/10.1016/j.cognition.2018.06.018}.
\item
  Falk, S., \& Kello, C. T. (2017). Hierarchical organization in the temporal structure of infant-direct speech and song. \emph{Cognition, 163}, 80--86. \url{https://doi.org/10.1016/j.cognition.2017.02.017}
\item
  Georg Northoff, Andrea Buccellato, Federico Zilio, Connecting brain and mind through temporo-spatial dynamics: Towards a theory of common currency, Physics of Life Reviews, Volume 52, 2025, Pages 29-43, ISSN 1571-0645, \url{https://doi.org/10.1016/j.plrev.2024.11.012}.
\item
  Northoff, G., \&\,Scalabrini, A. (2021, October 14). ``Project for a Spatiotemporal Neuroscience''~-- Brain and Psyche share their topography and dynamic. \emph{Frontiers in Psychology, 12}, Article 717402. \url{https://doi.org/10.3389/fpsyg.2021.717402}
\item
  Ao, Y., Klar, P., Catal, Y., Wang, Y., \& Northoff, G. (2025). Author Correction: Infra-slow scale-free dynamics modulate the connection of neural and behavioral variability during attention. \emph{Communications biology, 8}(1), 1205. \url{https://doi.org/10.1038/s42003-025-08603-w}
\item
  Wolman, A., Çatal, Y., Wolff, A., et al. (2023). Intrinsic neural timescales mediate the cognitive bias of self -- temporal integration as key mechanism. \emph{NeuroImage, 268}, 119896. \url{https://doi.org/10.1016/j.neuroimage.2023.119896}
\item
  Wang, T. and Hirst, G. (2011). Refining the Notions of Depth and Density in WordNet-based Semantic Similarity Measures. In \emph{Proceedings of the 2011 Conference on Empirical Methods in Natural Language Processing} (pp. 1003--1011). Association for Computational Linguistics.
\item
  Kumar, S., Klar, P., Çatal, Y., Chang, H. J., Pulvermüller, F., \& Northoff, G. (2025). From Speech Semantics to Brain Activity-Timescales Are Key in Their Information Transfer.~\emph{Human brain mapping},~\emph{46}(16), e70379. \url{https://doi.org/10.1002/hbm.70379}
\item
  \emph{Robyn Speer. (2022). rspeer/wordfreq: v3.0 (v3.0.2). Zenodo.~\url{https://doi.org/10.5281/zenodo.7199437}}
\item
  Dentella, V., Günther, F., Murphy, E.~\emph{et al.}~Testing AI on language comprehension tasks reveals insensitivity to underlying meaning.~\emph{Sci Rep}~\textbf{14}, 28083 (2024). \url{https://doi.org/10.1038/s41598-024-79531-8}
\item
  Frisby, S.L., Halai, A.D., Cox, C.R., Lambon Ralph, M.A., \& Rogers, T.T. (2023). Decoding semantic representations in mind and brain. \emph{Trends in Cognitive Sciences} 27(3): 258-281.
\item
  Murphy, E. (2025). ROSE: A universal neural grammar. \emph{Cognitive Neuroscience} 16(1-4): 49-80.
\item
  He, R., Palominos, C., Zhang, H., Alonso-Sánchez, M. F., Palaniyappan, L., \& Hinzen, W. (2024). Navigating the semantic space: Unraveling the structure of meaning in psychosis using different computational language models. \emph{Psychiatry Research}, \emph{333}, 115752. \url{https://doi.org/10.1016/j.psychres.2024.115752}
\item
  Futrell, R. (2024). An information-theoretic account of availability effects in language production. \emph{Topics in Cognitive Science, 16}(1), 38--53. \url{https://doi.org/10.1111/tops.12716}.
\item
  Jaeger, T. F. (2010). Redundancy and reduction: Speakers manage syntactic information density. \emph{Cognitive Psychology}, \emph{61}(1), 23--62. \url{https://doi.org/10.1016/j.cogpsych.2010.02.002}
\item
  \textbf{Aylett, M. (1999).} Stochastic suprasegmentals: Relationships between redundancy, prosodic structure and syllabic duration. \emph{Proceedings of ICPhS 1999}, 289--292.
\item
  \textbf{Aylett, M., \& Turk, A. (2004).} The Smooth Signal Redundancy Hypothesis: A functional explanation for relationships between redundancy, prosodic prominence, and duration in spontaneous speech. \emph{Language and Speech}, 47(1), 31--56. \url{https://doi.org/10.1177/00238309040470010201}
\item
  \textbf{Aylett, M., \& Turk, A. (2006).} Language redundancy predicts syllabic duration and the spectral characteristics of vocalic syllable nuclei. \emph{Journal of the Acoustical Society of America}, 119(5), 3048--3058. \url{https://doi.org/10.1121/1.2188331}
\item
  \textbf{Bell, A., Jurafsky, D., Fosler-Lussier, E., Girand, C., Gregory, M., \& Gildea, D. (2003).} Effects of disfluencies, predictability, and utterance position on word form variation in English conversation. \emph{Journal of the Acoustical Society of America}, 113(2), 1001--1024. \url{https://doi.org/10.1121/1.1534836}
\item
  \textbf{Genzel, D., \& Charniak, E. (2002).} Entropy rate constancy in text. In \emph{Proceedings of the 40th Annual Meeting of the Association for Computational Linguistics} (pp. 199--206). Association for Computational Linguistics. \url{https://doi.org/10.3115/1073083.1073117}
\item
  \textbf{Giulianelli, M., \& Fernández, R. (2021).} Analysing human strategies of information transmission as a function of discourse context. In \emph{Proceedings of the 25th Conference on Computational Natural Language Learning} (pp. 647--660). Association for Computational Linguistics. \url{https://doi.org/10.18653/v1/2021.conll-1.50}
\item
  \textbf{Giulianelli, M., Sinclair, A., \& Fernández, R. (2021).} Is information density uniform in task-oriented dialogues? In \emph{Proceedings of the 2021 Conference on Empirical Methods in Natural Language Processing} (pp. 8271--8283). Association for Computational Linguistics. \url{https://doi.org/10.18653/v1/2021.emnlp-main.652}
\item
  Ou, Y., Wang, Y., Xu, Y., \& Buschmeier, H. (2025). \emph{Identifying the Periodicity of Information in Natural Language} (arXiv:2510.27241). arXiv. \url{https://doi.org/10.48550/arXiv.2510.27241}
\item
  \textbf{Tsipidi, E., Nowak, F., Cotterell, R., Wilcox, E., Giulianelli, M., \& Warstadt, A. (2024).} Surprise! Uniform Information Density isn' t the whole story: Predicting surprisal contours in long-form discourse. In \emph{Proceedings of the 2024 Conference on Empirical Methods in Natural Language Processing} (pp. 18820--18836). Association for Computational Linguistics. \url{https://doi.org/10.18653/v1/2024.emnlp-main.1047}
\item
  \textbf{Tsipidi, E., Kiegeland, S., Nowak, F., Xu, T., Wilcox, E., Warstadt, A., Cotterell, R., \& Giulianelli, M. (2025).} The harmonic structure of information contours. In \emph{Proceedings of the 63rd Annual Meeting of the Association for Computational Linguistics} (pp. 31636--31659). Association for Computational Linguistics. \url{https://doi.org/10.18653/v1/2025.acl-long.1527}
\item
  \textbf{Xu, Y., \& Reitter, D. (2018).} Information density converges in dialogue: Towards an information-theoretic model. \emph{Cognition}, 170, 147--163. \url{https://doi.org/10.1016/j.cognition.2017.09.018}
\item
  Xu, Y., Wang, Y., An, H., Liu, Z., \& Li, Y. (2024). Detecting Subtle Differences between Human and Model Languages Using Spectrum of Relative Likelihood. In Y. Al-Onaizan, M. Bansal, \& Y.-N. Chen (Eds.), \emph{Proceedings of the 2024 Conference on Empirical Methods in Natural Language Processing} (pp. 10108--10121). Association for Computational Linguistics. \url{https://doi.org/10.18653/v1/2024.emnlp-main.564}
\item
  Yang, Z., Yuan, Y., Xu, Y., Zhan, S., Bai, H., \& Chen, K. (2023). \emph{FACE: Evaluating Natural Language Generation with Fourier Analysis of Cross-Entropy} (arXiv:2305.10307). arXiv. \url{https://doi.org/10.48550/arXiv.2305.10307}
\item
  \textbf{Brysbaert, M., Warriner, A. B., \& Kuperman, V. (2014).} Concreteness ratings for 40 thousand generally known English word lemmas. \emph{Behavior Research Methods}, 46(3), 904--911. \url{https://doi.org/10.3758/s13428-013-0403-5}
\item
  \textbf{Futrell, R., Gibson, E., \& Levy, R. P. (2020).} Lossy-context surprisal: An information-theoretic model of memory effects in sentence processing. \emph{Cognitive Science}, 44(3), e12814. \url{https://doi.org/10.1111/cogs.12814}
\item
  \textbf{Futrell, R. (2023).} Information-theoretic principles in incremental language production. \emph{Proceedings of the National Academy of Sciences}, 120(39), e2220593120. \url{https://doi.org/10.1073/pnas.2220593120}
\item
  \textbf{Hahn, M., Degen, J., \& Futrell, R. (2021).} Modeling word and morpheme order in natural language as an efficient trade-off of memory and surprisal. \emph{Psychological Review}, 128(4), 726--756. \url{https://doi.org/10.1037/rev0000269}
\item
  \textbf{Hahn, M., Futrell, R., Levy, R., \& Gibson, E. (2022).} A resource-rational model of human processing of recursive linguistic structure. \emph{Proceedings of the National Academy of Sciences}, 119(43), e2122602119. \url{https://doi.org/10.1073/pnas.2122602119}
\item
  \textbf{Hearst, M. A. (1997).} TextTiling: Segmenting text into multi-paragraph subtopic passages. \emph{Computational Linguistics}, 23(1), 33--64.
\item
  \textbf{Morris, J., \& Hirst, G. (1991).} Lexical cohesion computed by thesaural relations as an indicator of the structure of text. \emph{Computational Linguistics}, 17(1), 21--48.
\item
  \textbf{Stegemann-Philipps, C., Butz, M. V., Winkler, S., \& Achimova, A. (2021).} Speakers use more informative referring expressions to describe surprising events. \emph{Proceedings of the Annual Meeting of the Cognitive Science Society}, 43. \url{https://escholarship.org/uc/item/7374p5xq}
\item
  Gay, M., Haley, C., Giulianelli, M., \& Ponti, E. (2026). Is Information Density Uniform when Utterances are Grounded on Perception and Discourse? In V. Demberg, K. Inui, \& L. Marquez (Eds.), Proceedings of the 19th Conference of the European Chapter of the Association for Computational Linguistics (Volume 1: Long Papers) (pp. 3825--3853). Association for Computational Linguistics. \url{https://doi.org/10.18653/v1/2026.eacl-long.178}
\item
  Muraki, E. J., \& Pexman, P. M. (2026). Distinguishing abstraction from abstractness: Specificity norms for 8,500 English words. \emph{Behavior Research Methods}, \emph{58}(2), 60. \url{https://doi.org/10.3758/s13428-026-02949-7}
\item
  Puccetti, G., Esuli, A., \& Bolognesi, M. (2025). Wordnet and Word Ladders: Climbing the abstraction taxonomy with LLMs. In C. Zanchi, L. Brigada Villa, E. Biagetti, A. Rademaker, F. Bond, \& G. Rigau (Eds.), \emph{Proceedings of the 13th Global Wordnet Conference} (pp. 51--65). Global Wordnet Association. \url{https://doi.org/10.18653/v1/2025.gwc-1.7}
\item
  John P. McCrae, Haotian Zhu, Fei Xia, Al Waskow, and Kexin Gao. 2025.~\href{https://aclanthology.org/2025.gwc-1.16/}{Remedying Gender Bias in Open English Wordnet}. In~\emph{Proceedings of the 13th Global Wordnet Conference}, pages 133--141, Pavia, Italy. Global Wordnet Association.
\item
  Tily, H. and Piantadosi, S. T. (2009). Refer efficiently: Use less informative expressions for more predictable meanings. In Proceedings of the Workshop on the Production of Referring Expressions: Bridging the Gap Between Computational and Empirical Approaches to Reference. \href{https://colala.berkeley.edu/papers/tily09cogsci.pdf}{http://colala.berkeley.edu/papers/tily09cogsci.pdf}
\item
  Ravelli, A. A., Bolognesi, M. M., \& Caselli, T. (2025). Specificity ratings for English data. \emph{Cognitive Processing}, \emph{26}(2), 283--302. \url{https://doi.org/10.1007/s10339-024-01239-4}
\end{enumerate}

\end{document}